\pdfoutput=1

\documentclass[11pt]{article}

\usepackage{emnlp2021}

\usepackage{times}
\usepackage{latexsym}

\usepackage[T1]{fontenc}

\usepackage[utf8]{inputenc}

\usepackage{microtype}

\title{COVR: A Test-Bed for Visually Grounded Compositional Generalization with Real Images}

\author{\makecell{Ben Bogin$^{1}$ ~~~~~~~ Shivanshu Gupta$^{2}$   ~~~~~ Matt Gardner$^{3}$~~~~~ Jonathan Berant$^{1,3}$ } \\ 
$^{1}$Tel-Aviv University \hspace{4mm} $^{2}$University of California Irvine  \hspace{4mm}  $^{3}$Allen Institute for AI \\
\texttt{\makecell{\{ben.bogin,joberant\}@cs.tau.ac.il, shivag5@uci.edu,\\mattg@allenai.org\\
}}}

\usepackage{graphicx}
\usepackage{amssymb}
\usepackage{amsfonts}
\usepackage{amsmath}
\usepackage{svg}
\usepackage{algorithm}
\usepackage[noend]{algpseudocode}
\usepackage{makecell}
\usepackage{booktabs}
\usepackage{rotating}
\usepackage{enumitem}
\usepackage{pifont}
\usepackage[normalem]{ulem}
\usepackage{arydshln}

\usepackage{progressbar}

\newcommand\jb[1]{\textcolor{red}{[JB: #1]}}
  
\newif\ifcomments
\commentstrue
\ifcomments
    \providecommand{\bb}[1]{{\protect\color{olive}{[BB: #1]}}}
    \providecommand{\jb}[1]{{\protect\color{red}{[JB: #1]}}}
    \providecommand{\matt}[1]{{\protect\color{teal}{[MG: #1]}}}
    \providecommand\sanjay[1]{{\protect\color{magenta}{[Sanjay: #1]}}}
    \providecommand\sg[1]{{\protect\color{cyan}{[SG: #1]}}}
\else
    \providecommand{\bb}[1]{}
    \providecommand{\jb}[1]{}
    \providecommand{\matt}[1]{}
    \providecommand{\sanjay}[1]{}
    \providecommand{\sg}[1]{}
\fi

\newcommand{\visgroc}{\textsc{COVR}}
\newcommand{\lxmerttext}{\textsc{VB\textsubscript{\textsc{Text}}}}
\newcommand{\lxmertiid}{\textsc{VB}\textsubscript{iid}}

\definecolor{beaublue}{rgb}{0.74, 0.83, 0.9}
\definecolor{babypink}{rgb}{0.96, 0.76, 0.76}
\definecolor{bananamania}{rgb}{0.98, 0.91, 0.71}

\definecolor{applegreen}{rgb}{0.55, 0.71, 0.0}
\definecolor{apricot}{rgb}{0.98, 0.81, 0.69}
\definecolor{brightlavender}{rgb}{0.75, 0.58, 0.89}
\definecolor{brass}{rgb}{0.71, 0.65, 0.26}

\definecolor{buff}{rgb}{0.94, 0.86, 0.51}
\definecolor{caribbeangreen}{rgb}{0.0, 0.8, 0.6}
\definecolor{celadon}{rgb}{0.67, 0.88, 0.69}

\definecolor{blizzardblue}{rgb}{0.67, 0.9, 0.93}
\definecolor{bisque}{rgb}{1.0, 0.89, 0.77}

\begin{document}
\maketitle
\begin{abstract}
While interest in models that generalize at test time to new compositions has risen in recent years, benchmarks in the visually-grounded domain have thus far been restricted to synthetic images. In this work, we propose \visgroc{}, a new test-bed for visually-grounded compositional generalization with real images. To create \visgroc{}, we use real images annotated with scene graphs, and propose an almost fully automatic procedure for generating question-answer pairs along with a set of context images. \visgroc{} focuses on questions that require complex reasoning, including higher-order operations such as quantification and aggregation. Due to the automatic generation process, \visgroc{} facilitates the creation of compositional splits, where models at test time need to generalize to new concepts and compositions in a zero- or few-shot setting. 
We construct compositional splits using \visgroc{} and demonstrate a myriad of cases where state-of-the-art pre-trained language-and-vision models struggle to compositionally generalize. 
\end{abstract}

\section{Introduction}

Models for natural language understanding (NLU) have exhibited remarkable generalization abilities on many tasks, when the training and test data are sampled from the same distribution.
But such models still lag far behind humans when asked to generalize to an unseen combination of known concepts, and struggle to learn concepts for which only few examples are provided \cite{finegan-dollak-etal-2018-improving,Bahdanau2019CLOSUREAS}. Humans, conversely, do this effortlessly: for example, once humans learn the meaning of the quantifier \emph{``all''}, they can easily understand the utterance 
\emph{``all cheetahs have spots''} if they know what \emph{``cheetahs''} and \emph{``spots''} mean \cite{Chomsky1957-CHOSS-2,montague1970aug,fodor-1988}. This ability, termed \emph{compositional generalization}, is crucial for building models that generalize to new settings \cite{Lake2018BuildingMT}.

\begin{figure}
  \centering
  \includegraphics[width=0.9\linewidth]{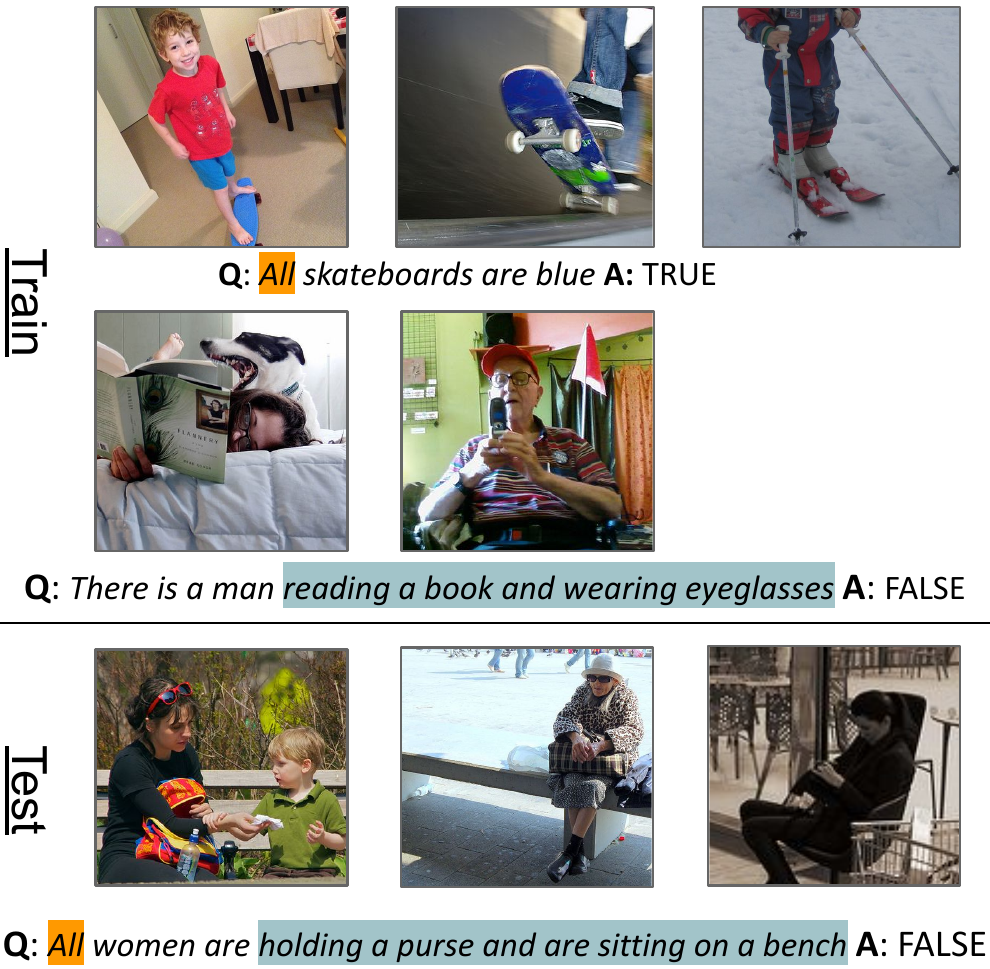}
  \caption{Compositional generalization in the VQA setup. \visgroc{} enables the creation of compositional splits such as the one depicted here, where quantification appears with conjunction only in the test set.}
  \label{fig:viscompgen-example}
\end{figure}

\begin{figure*}
  \centering
  \includegraphics[width=\linewidth]{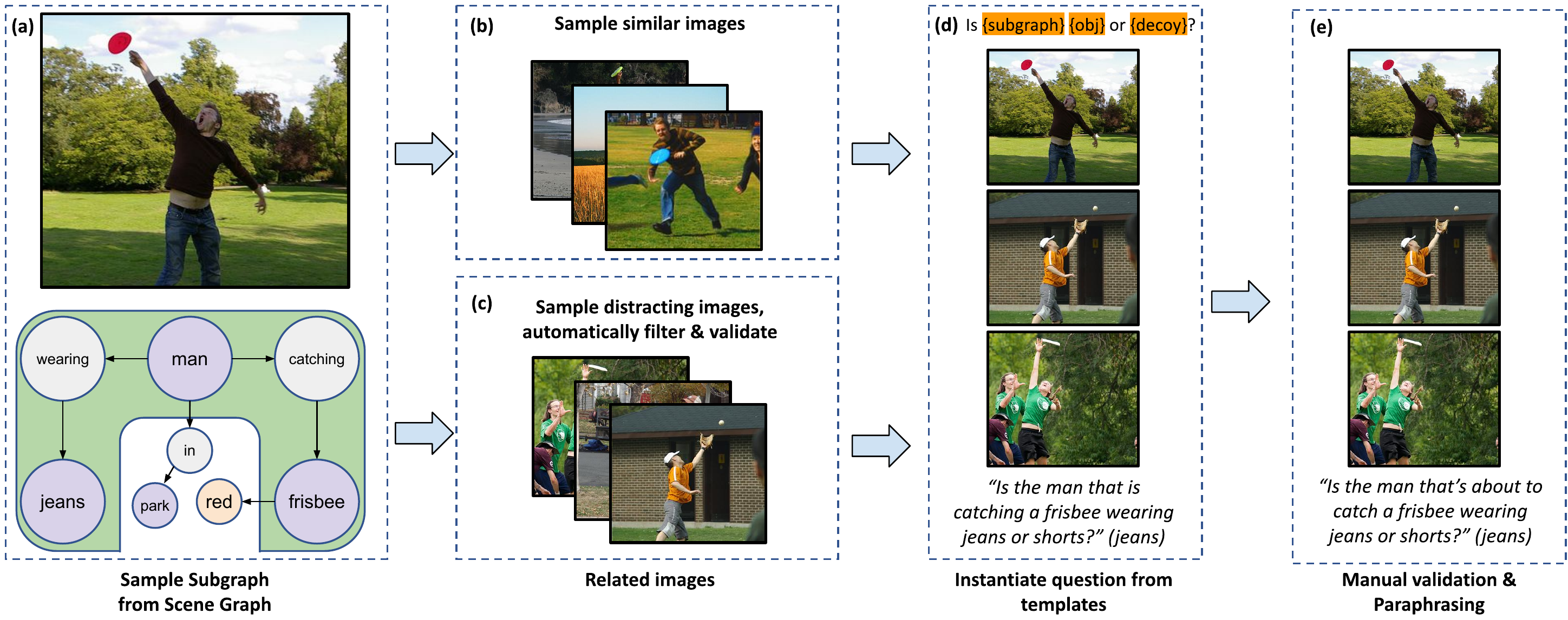}
  \caption{An overview of the dataset creation process.}
  \label{fig:creation}
\end{figure*}

In recent years, multiple benchmarks have been created, illustrating that current NLU models \emph{fail} to generalize to new compositions.
However, these benchmarks focused on semantic parsing, the task of mapping natural language utterances to logical forms \cite{Lake2018GeneralizationWS,kim-linzen-2020-cogs,keysers2020measuring}.
Visual question answering (VQA) is arguably a harder task from the perspective of compositional generalization, since the model needs to learn to compositionally ``execute'' the meaning of the question over images, without being exposed to an explicit meaning representation. For instance, in Fig.~\ref{fig:viscompgen-example}, a model should learn the meaning of the quantifier \emph{``all'} from the first example, and the meaning of a conjunction of clauses from the second example, and then execute both operations compositionally at test time.

Existing VQA datasets for testing compositional generalization typically use  synthetic images and contain a limited number of visual concepts and reasoning operators \cite{Bahdanau2019CLOSUREAS,bahdanau2018systematic,Ruis2020ABF}, or focus on generalization to unseen \textit{lexical} constructions rather than unseen reasoning skills \cite{hudson2019neuralstate,Agrawal2017CVQAAC}.
Other datasets such as GQA \cite{hudson2018compositional} use real images with synthetic questions, but lack logical operators available in natural language datasets, such as quantifiers and aggregations, and contain ``reasoning shortcuts'', due to a lack of challenging image distractors \cite{Chen_2020_CVPR}. 

In this work, we present \visgroc{} (\textbf{CO}mpositional \textbf{V}isual \textbf{R}easoning), a test-bed for visually-grounded compositional generalization with \emph{real} images. We propose a process for automatically generating complex questions over sets of images (Fig.~\ref{fig:creation}), where each example is annotated with the program corresponding to its meaning. We take images annotated with scene graphs (Fig.~\ref{fig:creation}a) from GQA, Visual Genome \cite{krishnavisualgenome}, and imSitu \cite{yatskar2016}, automatically collect both similar and distracting images for each example, and filter incorrect examples due to errors in the source scene graphs (Fig.~\ref{fig:creation}b,c). We then use a template-based grammar to generate a rich set of complex questions that contain multi-step reasoning and higher-order operations on multiple images (\ref{fig:creation}d).
To further enhance the quality of the dataset, we manually validate the correctness of development and test examples through crowdsourcing and paraphrase the automatically generated questions into fluent English for a subset of the automatically generated dataset (\ref{fig:creation}e). \visgroc{} contains 262k examples based on $\sim$89k images, with 13.9k of the questions manually validated and paraphrased.

Our automatic generation process allows for the easy construction of \emph{compositional data splits}, where models must generalize to new compositions, and is easily extendable with new templates and splits. We explore both the zero-shot setting, where models must generalize to new compositions, and the few-shot setting, where models need to learn new constructs from a small number of examples.

We evaluate state-of-the-art pre-trained models on a wide range of compositional splits, and expose generalization weaknesses in 10 out of the 21 setups, where the \emph{generalization score} we define is low (0\%-70\%). Moreover, results show it is not trivial to characterize the conditions under which generalization occurs, and we conjecture generalization is harder when it requires that the model learn to combine complex/large structures.
We encourage the community to use \visgroc{} to further explore compositional splits and investigate visually-grounded compositional generalization.\footnote{The dataset and our codebase can be found at \url{https://covr-dataset.github.io}.}
\section{Related Work}

Prior work on grounded compositional generalization has typically tested generalization in terms of \textit{lexical} and \textit{syntactic} constructions, while we focus on compositions in the space of \textit{meaning} by testing unseen program compositions. For example, the ``structure split'' in \citet{hudson2019neuralstate} splits examples based on their surface form. As a result, identical programs  are found in both the training and the test splits. The ``content split'' from the same work tests generalization to unseen lexical concepts. This is a different kind of skill than the one we address, where we assume the model sees all required concepts during training. C-VQA \cite{Agrawal2017CVQAAC} splits are based only on question-answer pairs and not on the question meaning. CLEVR-CoGenT \cite{johnson2017clevr} tests for unseen combinations of attributes, thus focuses more on visual generalization. Other datasets that do split samples based on their programs \cite{Bahdanau2019CLOSUREAS,bahdanau2018systematic,Ruis2020ABF} are using synthetic images with a small set of entities and relations ($\leq$ 20). 

GQA \cite{hudson2018compositional} uses real images with synthetic questions, which in theory could be used to create compositional splits. However, our work uses multiple images which allows testing reasoning operations over sets and not only over entities, such as counting, quantification, comparisons, etc. Thus, the space of possible compositional splits in COVR is much larger than GQA. If we anonymize lexical items in GQA programs (i.e., replace them with a single placeholder), the number of distinct programs (79) is too low to create a rich set of splits where all operators appear in the training set. In contrast, COVR contains 640 anonymized programs, allowing us to create a large number of splits. Moreover, our process for finding high-quality distracting images mitigates issues with reasoning shortcuts, where models solve questions due to a lack of challenging image distractors \cite{Chen_2020_CVPR}. Finally, other VQA datasets with real images and questions that were generated by humans \cite{suhr-etal-2019-corpus,VQA} do not include a meaning representation for questions and thus cannot be easily used to create compositional splits.

\section{Dataset Creation}

The goal of \visgroc{} is to facilitate the creation of VQA compositional splits, with questions that require a high degree of compositionality on both the textual and visual input.

\paragraph{Task definition} 
Examples in \visgroc{} are $(q, \mathcal{I}, a)$ triples, where $q$ is a complex question, $\mathcal{I}$ is a set of images, and $a$ the expected answer. Unlike most visually-grounded datasets , which contain 1-2 images, each example in \visgroc{} contains up to 5 images. This allows us to  (a) generate questions with higher-order operators,
and (b) detect good distracting images. Also, questions are annotated with programs corresponding to their meaning, which enables creating compositional splits.

\paragraph{High-level overview} Fig.~\ref{fig:creation} provides an overview of the data generation process. Given an image and its annotated scene graph describing the image objects and their relations, we iterate through a set of subgraphs. For example, 
in Fig.~\ref{fig:creation}a the subgraph corresponds to 
\emph{``a man is catching a frisbee and wearing jeans''}.
Next, we sample images with related subgraphs: (a) images that contain the same subgraph (Fig.~\ref{fig:creation}b), and (b) images that contain a similar subgraph, to act as distracting images (e.g., images with a man catching a ball or a woman catching a frisbee, Fig.~\ref{fig:creation}c). To ensure the quality of the distracting images, we propose models for automatic filtering and validation (\S\ref{subsec:distracing}). Next (Fig.~\ref{fig:creation}d), we instantiate questions from a template-based grammar by filling slots with values computed from the selected subgraphs, and automatically obtain a program for each question.
Last, we balance the answers and question types, and use crowdsourcing to manually validate and provide fluent paraphrases for the evaluation set and a subset of the training set (Fig.~\ref{fig:creation}e).

\subsection{Extracting Subgraphs}
\label{subsec:generating}

\begin{figure}
  \centering
  \includegraphics[width=0.67\linewidth]{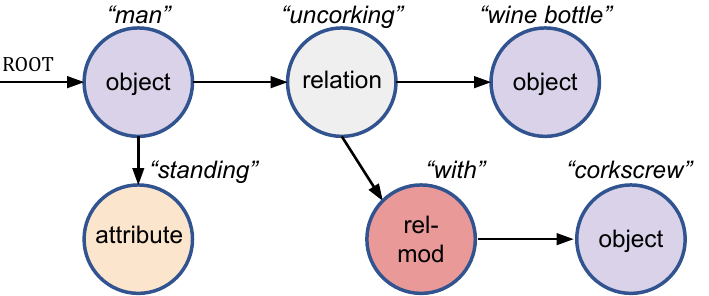}
  \caption{An example subgraph that shows all supported nodes, referring to \emph{``standing man uncorking a wine bottle with a corkscrew''}. The types of the nodes are inside the circle, and their name above it.}
  \label{fig:subgraph}
\end{figure}

A scene graph describes objects in an image, their attributes and relations.
We use existing datasets with annotated scene graphs, specifically imSitu \cite{yatskar2016} and GQA \cite{hudson2018compositional}, which contains a clean version of Visual Genome's human-annotated scene graphs \cite{krishnavisualgenome}. While Visual Genome's scenes have more detailed annotations, imSitu has a more diverse set of relations, such as \emph{``person measuring the length of the wall with a tape''}, which also introduces ternary relations  (\emph{``uncorking''}, in Fig.~\ref{fig:subgraph}).

Given a scene graph, we extract all of its \emph{subgraphs} using the rules described next. A subgraph is a directed graph with the following node types: \texttt{object}, \texttt{attribute}, \texttt{relation} and \texttt{rel-mod} (relation modifier), where every node has a name which describes it (e.g. \emph{``wine''}). Fig.~\ref{fig:subgraph} shows an example subgraph. A valid subgraph has a root \texttt{object} node, and has the following structure: Every \texttt{object} has an outgoing edge to $\leq 2$ \texttt{relation} nodes and $\leq 1$ \texttt{attribute} node.
Every \texttt{relation} node has an outgoing edge to exactly one \texttt{object} node and optionally multiple \texttt{rel-mod} nodes. Every \texttt{rel-mod} node has an outgoing edge to exactly one \texttt{object} node.
The depth of the subgraph is constrained such that every path from the root to a leaf goes through at most two \texttt{relation} nodes.
For brevity, we refer to subgraphs with a textual description (\emph{``standing man uncorking a wine bottle with a corkscrew''}).

\subsection{Finding Related Images}
\label{subsec:distracing}
Given a subgraph $g$,
we pick candidate \emph{context images} that will be part of the example images $\mathcal{I}$. We want to only pick \emph{related} images, i.e., images that share sub-structure with $g$. Images that contain $g$ in their scene graph will be used to generate counting or quantification questions. Images with different but similar subgraphs to $g$ will be useful as image distractors, which are important to avoid ``reasoning shortcuts'' \cite{cirik-etal-2018-visual,AgrawalBPK18,chen2020counterfactual,Chen_2020_CVPR,bitton2020automatic,kervadec2021transferable}.
For example, for the question in Fig.~\ref{fig:creation}e, it is not necessary to perform all reasoning steps if there is only one man in all images in $\mathcal{I}$. 

\begin{table}[!t]
\centering
\scriptsize
\begin{tabular}{p{1.7cm}p{5.05cm}}
\toprule
{\bf Type} & {\bf Example} \\
 \midrule
  \textsc{VerifyAttr} & Is \underline{the sink that is below a towel} \underline{white}? \\
  \textsc{ChooseAttr} & Is \underline{the man that is wearing a jersey} \underline{running} or \underline{looking up}? \\
  \textsc{QueryAttr} & What is the \underline{color} of \underline{the cat that is on a black} \underline{floor}? \\
  \textsc{CompareCount} & There are \underline{more} \underline{coffee tables that are in living} \underline{room} than \underline{couches that are in living room} \\
  \textsc{Count} & How many \underline{people are erasing a mark from a} \underline{paper}? \\
  \textsc{VerifyCount} & There is \underline{at most} \underline{1} \underline{cup that is behinda man} \underline{that is} \underline{wearing a jacket} \\
  \textsc{CountGroupBy} & How many images contain \underline{exactly} \underline{2} \underline{men that are in water}? \\
  \textsc{VerifyCount-GroupBy} & There is \underline{at least 1} image that contain exactly \underline{2} \underline{women that are carrying surfboard} \\
  \textsc{VerifyLogic} & Are there \underline{both} \underline{people that are packing a tea} \underline{into a mason jar} \underline{and} \underline{people that are packing a} \underline{salt into a bag}? \\
  \textsc{VerifyQuant} & \underline{No} \underline{boys with large trees behind them} are \underline{wearing} \underline{jeans} \\
  \textsc{VerifyQuant-}\textsc{Attr} & Do all \underline{dogs that are on a bed} have the same \underline{color}? \\
  \textsc{ChooseObject} & \underline{The woman that is wearing dress} is \underline{carrying} \underline{a bottle} or \underline{a purse}? \\
  \textsc{QueryObject} & What is \underline{the woman that is wearing glasses} \underline{holding}? \\
  \textsc{VerifySame-Attr} & Does \underline{the pillow that is on a bed} and \underline{the pillow} \underline{that is on a couch} have the same \underline{color}? \\
  \textsc{ChooseRel} & Is \underline{the sitting man} \underline{holding} \underline{a hat} or \underline{wearing} it? \\
 \bottomrule
\end{tabular}
\caption{A list of all question templates with examples. Template slots are underlined.}
\label{tab:templates_all}
\end{table}

\begin{table*}[!t]
\centering
\scriptsize
\begin{tabular}{p{1.0cm}p{2.9cm}p{3.2cm}p{2.8cm}p{4.1cm}}
\toprule
{\bf Type} & {\bf Preconditions} & {\bf Example subgraph(s)} & {\bf Template} & {\bf Example output}\\
 \midrule
  \textsc{Verify}-\textsc{Attr} & (1) $g$'s root has attribute \newline (2) Distracting images have 2 different nodes, one of which is the root's attribute &  \colorbox{babypink}{white} \colorbox{beaublue}{sink below towel} & 
  Is the \colorbox{beaublue}{\textsc{G-NoAttribute}}, \colorbox{babypink}{\textsc{Attribute}}? &
  Is the \colorbox{beaublue}{sink that is below a towel}, \colorbox{babypink}{white}? \\
  \midrule
  \textsc{Choose-}\textsc{Object} & (1) Distracting images have 2 different nodes, one of which is the object & \colorbox{applegreen}{woman lighting a cigar on fire} \colorbox{apricot}{using} \colorbox{brightlavender}{a candle} & 
  \colorbox{applegreen}{\textsc{G-Subject}} is \colorbox{apricot}{\textsc{Rel}} \colorbox{brightlavender}{\textsc{Obj}} or \colorbox{brass}{\textsc{DecoyObj}}? &
  \colorbox{applegreen}{The woman that is lighting a cigar on fire} is \colorbox{apricot}{lighting it using} a \colorbox{brightlavender}{candle} or \colorbox{brass}{a lighter}? \\
  \midrule
  \textsc{Compare-}\textsc{Count} & (no conditions) & \colorbox{buff}{woman wearing dark blue jacket},\newline \colorbox{caribbeangreen}{woman wearing white jacket} & 
  There are \colorbox{celadon}{\textsc{Comparative}} \colorbox{buff}{\textsc{G}} than \colorbox{caribbeangreen}{\textsc{G2}} &
  There are \colorbox{celadon}{less} \colorbox{buff}{women that are wearing a dark blue jacket} than \colorbox{caribbeangreen}{women that are wearing a white jacket}
 \\
 \bottomrule
\end{tabular}
\caption{A subset of the question templates with their preconditions and examples for how questions are instantiated. Template slots are shown with a background color. See text for further explanation.}
\label{tab:templates}
\end{table*}

To find images with similar subgraphs, we define the \emph{edit distance} between two graphs to be the minimum number of operations required to transform one graph to another, where the only valid operation is to \emph{substitute} a node with another node that has the same type. A good distracting image will (a) contain a subgraph at edit distance $1$ or $2$, 
and (b) not contain the subgraph $g$ itself. For the question in Fig.~\ref{fig:creation}e, a good distracting image will contain, for example, \emph{``a man holding a frisbee wearing shorts''} or \emph{``a woman catching a frisbee wearing jeans''}. 
We extract related images by querying all scene graphs that exhibit the mentioned requirements using a graph-based database.\footnote{\url{https://neo4j.com/}}

\noindent
\textbf{Filtering overlapping distractors}
A drawback of the above method is that the edit distance between two subgraphs could be 1, but the two subgraphs might still semantically \emph{overlap}. For example, a subgraph \emph{``woman using phone''} is not a good distractor for \emph{``woman holding phone''} since \emph{``holding''} and \emph{``using''} are not mutually exclusive.
A similar issue arises with objects and attributes (e.g. \emph{``man''} and \emph{``person''}, \emph{``resting''} and \emph{``sitting''}).

We thus add a step when sampling distracting images to filter such cases. We define $m(x_1, x_2)$ to be the probability that node $x_1$ and $x_2$ are mutually exclusive. For example, for an object $x$, $m(x, \cdot)$ should return high probabilities for all objects except $x$, its synonyms, hypernyms, and hyponyms. When performing node substitutions (to compute the edit distance) we only consider nodes $x_1, x_2$ such that $m(x_1, x_2)>0.5$. To learn $m(x, \cdot)$, we fine-tune RoBERTa \cite{liu2019roberta} separately on nouns, attributes and relations, with a total of 6,366 manually-annotated word pairs, reaching accuracies of 94.4\%, 95.7\% and 82.5\% respectively. See App.~\ref{app:overlapping} for details.

\noindent
\textbf{Incomplete scene graphs}
A good distracting image should not contain the source subgraph $g$.
However, because scene graphs are often incomplete, na\"ive selection from scene graphs can yield a high error rate. To mitigate this issue, we fine-tune LXMERT \cite{tan-bansal-2019-lxmert}, a pre-trained multi-modal classifier, to recognize if a given subgraph is in an image or not. We do not assume that this classifier will be able to comprehend complex subgraphs, and only use it to recognize \emph{simple} subgraphs, that contain $\leq 2$ \texttt{object} nodes, and up to a single \texttt{relation} and \texttt{attribute}. We train the model on binary-labeled image-subgraph pairs, where we sample a simple subgraph $g_s$ and its image $I$ from the set of all subgraphs, and use $I$ as a positive example, and an image $I'$ that contains a subgraph $g_s'$ as a negative example, where $I'$ is a distracting image for $I$, according to the procedure described above. 
For example, for the subgraph \emph{``man wearing jeans''} the model will be trained to predict `True' for an image that contains it, and `False' for an image with \emph{``man wearing shorts''}, but not \emph{``man with jeans''}.

After training, we filter out candidate distracting images for the subgraph $g$ if the model outputs a score above a certain threshold $\tau$ for \emph{all} of the simple graphs in $g$.
We adjust $\tau$ such that the probability of correctly identifying missing subgraphs is above 95\%  according to a manually-annotated set of 441 examples. We also use our trained model to filter out cases where an object is annotated in the scene graph, but the image contains other instances of that object (e.g., if in an image with a crowd full of people, only a few are annotated). See App.~\ref{app:incomplete} for details.

\subsection{Template-based Question Generation}
Once we have a subgraph, an image, and a set of related images, we can generate questions. For a subgraph $g$ and its related images, we generate questions from a set of 15 manually-written templates (Table~\ref{tab:templates_all}), that include operators such as quantifiers, counting, and ``group by''. Further extending this list is trivial in our framework.
Each template contains slots, filled with values conditioned on the subgraph $g$ as described below.
Since we have 15 templates with multiple slots, and each slot can be filled with different types of subgraphs, we get a large set of possible question types and programs. Specifically, if we anonymize lexical terms (i.e. replace nouns, relations and attributes with a single placeholder) there are 640 different programs in total.

We define a set of \emph{preconditions} for each template (Table~\ref{tab:templates}) which specifies (a) the types of subgraphs that can instantiate each slot, and (b) the images that can be used as distractors. The first precondition type ensures that the template is supported by the subgraph---e.g., to instantiate a question that verifies an attribute (\textsc{VerifyAttr}), the root object must have an outgoing edge to an \texttt{attribute}. The second precondition type ensures that we have relevant distractors. For example, for \textsc{VerifyAttr}, we only consider distracting subgraphs if the attribute connected to their root is \emph{different} from the attribute of $g$'s root. In addition, the distracting subgraphs must have at least one more different node. Otherwise, we cannot refer to the object unambiguously: if we have a \emph{``white dog''} as a distracting image for a \emph{``black dog''}, the question \emph{``Is the dog black?''} will be ambiguous.

\begin{table}[]
\centering
\small
\begin{tabular}{lll}
\toprule
Measurement & Train & Dev.+Test \\
\midrule
\# total questions & 248.1k & 13.9k\\
\# unique questions & 122.0k & 7.6k \\
\# unique answers & 3666 & 1268 \\
\# unique images & 79.0k & 9.5k \\
\# unique anonymized programs & 635 & 291 \\
\# True/False (T/F) questions & 133.3k & 7.5k \\
\# ``X or Y'' questions & 50.0k & 2.9k\\
\# ``how many ...'' questions & 33.3k & 1.8k \\
\# Open questions & 31.5k & 1.7k\\
avg. \# question words (a/p) & 14.0 & 13.5/11.9 \\
avg. \# images per question & 4.4 & 3.4 \\
\bottomrule
\end{tabular}
\caption{Statistics for \visgroc{} for both the training and test sets (development and test combined). a/p stands for automatically-generated vs paraphrased.}
\label{tab:stats}
\end{table}

When a template, subgraph, and set of images satisfy the preconditions, we instantiate a question by filling the template slots. 
There are three types of slots: First, slots with the description of $g$ or a subset of its nodes. E.g., in \textsc{VerifyAttr} (Table~\ref{tab:templates}) we fill the slot \colorbox{beaublue}{\textsc{G-NoAttribute}} with the description of $g$ without the attribute node (\emph{``sink that is below a towel''}), and the slot \colorbox{babypink}{\textsc{Attribute}} with the name of that attribute (\emph{``white''}). In \textsc{ChooseObject} (second row), we fill the slots  \colorbox{applegreen}{\textsc{G-Subject}}, \colorbox{apricot}{\textsc{Rel}} and \colorbox{brightlavender}{\textsc{Obj}} with different subsets of $g$: \emph{``woman lighting a cigar on fire''}, \emph{``using''} and \emph{``candle''}.

The second slot type fills the description of a \emph{different} subgraph than $g$, or a subset of its nodes. In \textsc{CompareCount} (third row), we fill \colorbox{caribbeangreen}{\textsc{G2}} with the description of another subgraph, sampled from the distracting images. Similarly, in \textsc{ChooseObject}, we fill \colorbox{brass}{\textsc{DecoyObj}} with the node \emph{``lighter''}.
Last, some slots are filled from a closed set of words that describe reasoning operations: In \textsc{CompareCount}, we fill \colorbox{celadon}{\textsc{Comparative}} with one of \emph{``less''}, \emph{``more''} or \emph{``same number''}.

\begin{table*}[]
\centering 
\scriptsize
\begin{tabular}{p{3.5cm}p{6.3cm}p{4.8cm}}
\toprule
Test name & Training & Generalization \\ \midrule 

\textsc{Has-Quant-CompScope \&}\newline \textsc{Has-Quant-All} & No \colorbox{bisque}{man that is next to a horse} is standing  \newline \colorbox{blizzardblue}{All birds are black} & \colorbox{blizzardblue}{All} \colorbox{bisque}{computer mice that are on a} \colorbox{bisque}{mouse pad} \colorbox{blizzardblue}{are black}\\
\hdashline
\textsc{Has-Count} \& \textsc{Has-Attr} & \colorbox{blizzardblue}{There are 3} dogs \newline What is the \colorbox{bisque}{black} dog holding? & \colorbox{blizzardblue}{There are 3} \colorbox{bisque}{black} dogs\\
\hdashline

\textsc{Has-Count \& RM/V/C} & The horse is pulling people \colorbox{blizzardblue}{with a rope} or a leash? \newline \colorbox{bisque}{There are two} people next to a tree. & 
\colorbox{bisque}{There are two} children cleaning the path \colorbox{blizzardblue}{with a broom}\\
\hdashline

\textsc{\textsc{Lexical-X/Lexical-Y}}\newline (Lexical Split) & All men wear \colorbox{blizzardblue}{jeans}. \newline No women are \colorbox{bisque}{standing}. & 
No men wearing \colorbox{blizzardblue}{jeans} are \colorbox{bisque}{standing}.\\

\hline

\textsc{Has-SameAttr-Color} & What is the \colorbox{blizzardblue}{color} of X? \newline \colorbox{bisque}{Do all X have the same} material? & \colorbox{bisque}{Do all X have the same} \colorbox{blizzardblue}{color}?\\
\hdashline
\textsc{TPL-ChooseObject} & What is the \colorbox{blizzardblue}{man carrying}? \newline Is the man \colorbox{bisque}{X or Y}?  & The \colorbox{blizzardblue}{man is carrying} a \colorbox{bisque}{X or Y}?\\
\hdashline
\textsc{TPL-VerifyQuantAttr} &  
Does the sitting dog and the standing cat \colorbox{blizzardblue}{have the same color}? \newline
\colorbox{bisque}{All dogs are} standing
& Do \colorbox{bisque}{all dogs have} the \colorbox{blizzardblue}{same color}? \\
\hdashline
\textsc{TPL-VerifyAttr} &  
Is the table that is under donuts \colorbox{blizzardblue}{dark or tan}?
& Is the towel that is on a floor \colorbox{blizzardblue}{pink}? \\
\hdashline
\textsc{TPL-VerifyCount} $\cup$ \newline \textsc{TPL-VerifyCountGroupBy}  & 
How many \colorbox{blizzardblue}{images contain at least 2} men that are in water?
& 
There are \colorbox{blizzardblue}{at least 2} images that contain exactly 2 blankets that are on bed\\

\bottomrule
\end{tabular}
\caption{List of the zero-shot compositional splits. Top half shows splits where we hold out examples where two properties co-occur, bottom half shows splits 
where we hold out questions with a single property or a union of two (see text). Background colors highlight different reasoning steps that the model is trained or tested on.}
\label{tab:splits}
\end{table*}

Once slots are filled, we compute the corresponding program for the question. Each row in the program is an operator (such as \texttt{Find}, \texttt{Filter}, \texttt{All}) with a given set of arguments (e.g. \emph{``black''}) or dependencies (input from other operator). The list of operators and a sample program are in App.~\ref{app:programs}. 

\subsection{Quality Assurance and Analysis}
\label{sec:post}

We perform the generation process separately on the training and validation set of GQA and imSitu graph scenes, which yields 13 million and 1 million questions respectively. We split the latter set into \emph{development} and \emph{test} sets of equal size, making sure that no two examples with the same question text appear in both of them.

\noindent
\textbf{Balancing the dataset}
Since we generate a question for every valid combination of subgraph and template, the resulting set of questions possibly contains correlations between the language of the question and the answer, and has skewed distributions over answers, templates, and subgraph structures. To overcome the first issue, whenever possible, we create two examples with the same question, but a different set of images and answers (186,841/75.3\% questions appear with at least two different answers in the training set).
Then, to balance our dataset, we use a heuristic procedure that leads to a nearly uniform distribution over templates, and balances the distribution over answers and over the size and depth of subgraphs as much as possible (App.~\ref{app:balance}).
We provide statistics about our dataset in Table~\ref{tab:stats} and the answer distribution in App.~\ref{app:answers_dist}.

\noindent
\textbf{Manual Validation}
Examples thus far are generated automatically. 
To increase the evaluation set quality, we validate all development and test examples through crowdsourcing. Workers receive the question, images and answer, and reject it if they find it invalid, leading to a 17\% rejection rate (see App.~\ref{subapp:validation} for more details, including error analysis).

\noindent
\textbf{Paraphrasing}
Since questions are generated automatically, we ask workers to paraphrase 3990 examples from the training set, and all development and test questions into fluent English, while maintaining their meaning. 
Paraphrasing examples are given in App.~\ref{subapp:paraphrasing}.

Overall, after validation, \visgroc{} contains 262,069 training, 6,891 development and 7,024 test examples. See App.~\ref{app:examples} for examples from the validation set.

\section{Compositional Splits}

Our generation process lets us create questions and corresponding programs with a variety of reasoning operators. We now show how \visgroc{} can be used to generate challenging compositional splits.
We propose two setups: (1) \emph{Zero-shot}, where training questions provide examples for all required reasoning steps, but the test questions require a new \emph{composition} of these reasoning steps, and (2) \emph{Few-shot}, where the model only sees a small number of examples for a specific reasoning type \cite{Bahdanau2019CLOSUREAS,yin-etal-2021-compositional}.

Because each question is annotated with its program, we can define binary properties over the program and answer, where a property is a binary predicate that typically defines a reasoning type in the program. For example, the property \textsc{Has-Quant} is true \emph{iff} the program contains a quantifier, and \textsc{Has-Quant-None} \emph{iff} it contains the quantifier \textsc{None}. We can create any compositional split that is a function of such properties.
We list the types of properties used in this work in Table~\ref{tab:properties}.

All compositional splits are based on the original training/validation splits from Visual Genome and imSitu to guarantee that no image appears in both the training set and any of the test sets.
Splits are created simply by removing certain questions from the train and test splits. If we do not remove any question, we get an i.i.d split. 

\begin{table}[]
\centering
\scriptsize
\begin{tabular}{p{1.55cm}p{5.4cm}}
\toprule
Property & Description \\ \midrule 
\textsc{Has-X} & True if $p$ contains an operator of type $\text{X}$, $\text{X} \in \{$ \textsc{Quantifier} (\textsc{Quant}), \textsc{Comparative} (\textsc{Compar}), \textsc{GroupBy} (\textsc{Group}), \textsc{Number} (\textsc{Num}), \textsc{Attribute} (\textsc{Attr}), \textsc{SameAttr}$\}$. \\
\textsc{Has-X-Y} & Same as \textsc{Has-X}, where Y is a specific instance of X (e.g., \textsc{All} if X is \textsc{Quantifier}). \\
\textsc{Has-Quant-CompScope} & True if the quantifier's scope is ``complex'', i.e.,  includes an attribute or a relation.\\
\textsc{RM/V/C} & True if $g$'s structure contains either a \texttt{Rel-Mod} node, an \texttt{object} node with two outgoing edges to \texttt{relation} nodes (V-shape) or chain of more than a single relation (C). \\
\textsc{TPL-X} &  True if the question originated from the template X.\\
\textsc{Ans-X} &  True if answer is of type $\text{X} \in \{$ \textsc{Num, Attr, Noun}$\}$\\
\textsc{Lexical-X} & True if $g$ contains a node with the name X. \\
\bottomrule
\end{tabular}
\caption{List of the types of properties given a program $p$ and subgraph $g$ on which the question was based on. 
}
\label{tab:properties}
\vspace{-4pt}
\end{table}
\noindent
\textbf{Zero-shot}
We test if a model can answer questions where two properties co-occur, when during training it has only seen questions that have at most one of these properties. For a pair of properties, we filter from the training set all examples that have \emph{both} properties, and keep only such examples for the evaluation set.
For example, the split in the first row of Table~\ref{tab:splits} (top) shows a split of the two properties \textsc{Has-Quant-CompScope} and \textsc{Has-Quant}.
    
The zero-shot setup can also be used to test examples with a single property that is unseen during training (Table~\ref{tab:splits}, bottom), or a \emph{union} of two properties, assuming that the model has seen examples for all the reasoning steps that the unseen property requires. For example, in \textsc{TPL-ChooseObject} we test on a template that is entirely unseen during training, since we have the templates \textsc{ChooseAttr} and \textsc{VerifyObject}.

Another popular zero-shot test is the \emph{program split}. In a similar fashion to the compositional generalization tests in \citet{finegan-dollak-etal-2018-improving}, we randomly split programs after anonymizing the names of nodes, and hold out 20\% of the programs to test how models perform on program structures that were not seen during training. We also perform a \emph{lexical split}, where we hold out randomly selected pairs of node names (i.e., names of objects, relations or attributes) such that the model never sees any pair together in the same program during training. We create 3 random splits where we hold out a sample of 65 lexical pairs, making sure that each lexical \textit{term} appears at least 50 times in the training set.

\noindent
\textbf{Few-shot}
In this setup, we test if a model can handle examples with a given property, when it has only seen a small number $M$ of examples with this property during training. For a given property, we create this split by filtering from the original training set all examples that have this property, except for $M$ examples.
From the evaluation set, we keep only examples that have this property.

\begin{table}[]
\centering
\scriptsize
\begin{tabular}{lcccc}
\toprule
Model & \multicolumn{2}{c}{\visgroc{}} & \multicolumn{2}{c}{\visgroc{}-\textsc{Paraph.}} \\
      & \multicolumn{1}{c}{Development}        & \multicolumn{1}{c}{Test}        & \multicolumn{1}{c}{Development}            & \multicolumn{1}{c}{Test}            \\ \midrule 
\textsc{Maj}        & 26.4        & 26.8        & 26.4           & 26.8       \\
\textsc{MajTempl}   & 42.1        & 41.5        & 42.1           & 41.5        \\
\lxmerttext{}       & 45.6        & 44.8        & 40.9           & 39.9         \\
\lxmertiid{}        & 69.2        & 67.6        & 61.1           & 57.9          \\
\midrule
VB\textsubscript{\textsc{EasyDistractors}} & 56.4 & 56.9 & 50.2 & 49.0  \\
\midrule
Humans & - & - & - & 91.9 \\
\bottomrule
\end{tabular}
\caption{Results on both the generated and paraphrased versions of the development and test set for the i.i.d. split.}
\label{tab:results_iid}
\end{table}

\begin{table*}[]
\centering
\scriptsize
\begin{tabular}{lllllll}
\toprule
Split & Filtered & \lxmerttext{} & VB\textsubscript{250} & Gen. Score & VB\textsubscript{iid-size} & VB\textsubscript{iid} \\ \midrule
\textsc{Has-Quant} & 33.3k & 50.8 & 55.8 & \progressbar{0.18} & 78.1 & 80.5  \\
\textsc{Has-Quant-All} & 21.1k & 50.7 & 69.2 & \progressbar{0.76} & 75.2 & 77.6  \\
\textsc{Has-Quant-CompScope} & 22.8k & 50.8 & 60.4 & \progressbar{0.35} & 78.1 & 80.3  \\
\textsc{Has-Compar} & 16.7k & 54.3 & 57.5 & \progressbar{0.15} & 76.4 & 80.0  \\
\textsc{Has-Compar-More} & 7.6k & 53.8 & 79.8 & \progressbar{0.94} & 81.3 & 84.0  \\
\textsc{Has-GroupBy} & 33.3k & 38.1 & 56.5 & \progressbar{0.74} & 62.8 & 65.7  \\
\textsc{Has-Logic} & 16.7k & 50.4 & 69.3 & \progressbar{0.75} & 75.4 & 77.5  \\
\textsc{Has-Logic-And} & 8.8k & 50.4 & 74.8 & \progressbar{1.11} & 72.3 & 75.2  \\
\textsc{Has-Num-3} & 6.9k & 44.8 & 68.8 & \progressbar{0.92} & 70.8 & 71.9  \\
\textsc{Has-Num-3-Ans-3} & 12.1k & 26.2 & 25.8 & \progressbar[linecolor=red]{0} & 65.6 & 66.8  \\
\textsc{Ans-Num} & 33.3k & 26.4 & 32.2 & \progressbar{0.15} & 64.9 & 67.0  \\
\bottomrule
\end{tabular}
\caption{Non-paraphrased test results in the few-shot setup. `Filtered' shows the number of examples that were filtered out of the training set in each split.}
\label{tab:results_few_shot}
\end{table*}
\begin{table*}[]
\centering
\scriptsize
\begin{tabular}{m{6cm}lllllll}
\toprule
Split & Filtered & \lxmerttext{} & VB\textsubscript{0} & Gen. Score & VB\textsubscript{iid-size} & VB\textsubscript{iid} \\ \midrule
\textsc{Has-Quant-CompScope \& Has-Quant-All} & 12.9k & 50.8 & 57.7 & \progressbar{0.26} & 77.3 & 78.2  \\
\textsc{Has-Count \& Has-Attr} & 37.6k & 41.2 & 58.7 & \progressbar{0.82} & 62.6 & 63.6  \\
\textsc{Has-Count \& RM/V/C} & 36.9k & 40.5 & 74.1 & \progressbar{0.81} & 82.2 & 82.0  \\
\textsc{Has-SameAttr-Color} & 27.3k & 49.8 & 66.0 & \progressbar{0.76} & 71.2 & 72.6  \\
\textsc{TPL-ChooseObject} & 16.7k & 52.0 & 1.6 & \progressbar[linecolor=red]{0} & 62.6 & 66.4  \\
\textsc{TPL-VerifyQuantAttr} & 16.7k & 50.4 & 71.2 & \progressbar{0.78} & 76.9 & 80.3  \\
\textsc{TPL-VerifyAttr} & 16.7k & 49.6 & 0.0 & \progressbar[linecolor=red]{0} & 75.4 & 73.5  \\
\textsc{TPL-VerifyCount} $\cup$ \textsc{TPL-VerifyCountGroupBy} & 33.3k & 49.8 & 41.7 & \progressbar[linecolor=red]{0} & 77.6 & 79.8  \\
\midrule
Program Split & 48k $\pm$ 11k & 43.8 $\pm$ 4.5 & 49.5 $\pm$ 3.6 & \progressbar{0.32} & 61.5 $\pm$ 4.4 & 64.8 $\pm$ 4.7  \\
Lexical Split & 40k $\pm$ 3k & 47.5 $\pm$ 0.5 & 69.3 $\pm$ 1.4 & \progressbar{0.93} & 71.0 $\pm$ 0.4 & 73.7 $\pm$ 0.6  \\
\bottomrule
\end{tabular}
\caption{Non-paraphrased test results in the zero-shot setup. Filtered shows number of examples that were filtered out of the training set. A red rectangle under ``Gen. Score'' illustrates that VB\textsubscript{0} is lower than \lxmerttext{}. `\&` indicates holding out the intersection of two sets of questions, `$\cup$` indicates holding out the union of the two.}
\label{tab:results_zero_shot}
\vspace{-2pt}
\end{table*}

\section{Experiments}

\paragraph{Experimental Setup}
We consider the following baselines: (a) \textsc{Maj}, the majority answer in the training set, and (b) \textsc{MajTempl}:  an oracle-based baseline that assumes perfect knowledge of the template from which the question was generated, and predicts the majority answer for that template. For templates that include two possible answers (\emph{``candle or lighter''}), it randomly picks one. 

We use the Volta framework \cite{bugliarello-etal-2021-multimodal} to train and evaluate different pre-trained vision-and-language models. We use Volta's controlled setup models (that have a similar number of parameters and pre-training data) of VisualBERT \cite{li2019visualbert} and \textsc{ViLBERT} \cite{vilbert}. In this section we show results only for VisualBERT, and results for \textsc{ViLBERT} can be found in App. \ref{app:additional_results}, showing mostly similar scores.

A vision-and-language model provides a representation for question-image pairs, $(q,I)$.
We modify the implementation to accept $N$ images by running the model with $(q, I)$ as input for each image $I \in \mathcal{I}$, and then passing the $N$ computed representations through two transformer layers with a concatenated \texttt{[CLS]} token. We pass the \texttt{[CLS]} token representation through a classifier layer to predict the answer.
The classifier layer and the added transformer layers are randomly initialized, and all parameters are fine-tuned during training.

To estimate a lower bound on performance without any reasoning on images, we evaluate a text-only baseline \lxmerttext{} that only sees the input text (image representations are zeroed out).

For the compositional splits, we evaluate VisualBERT trained on the entire data (\lxmertiid), the text baseline (\lxmerttext{}), and the compositionally-trained models VB\textsubscript{250}, VB\textsubscript{0} for the few-shot ($M$=250) and zero-shot setups, respectively. To control the training size, we also evaluate VB\textsubscript{iid-size}, a model trained with a similar data size as the compositionally-trained model, by uniformly downsampling the training set.
All models are evaluated on the same subset of the development compositional split. To focus on the generalization gap, we define a generalization score (``Gen. Score'') that measures the proportion of the gap between \lxmerttext{} and our upper-bound, VB\textsubscript{iid--size}, that is closed by a model.
In all compositional splits, we train the models 8 epochs and early-stop using the subset of the development set that does not contain any of the compositional properties we test on \cite{Teney2020OnTV}.

\paragraph{Results} 
First, we show how models perform on paraphrased and automatically-generated questions in the i.i.d setup in Table~\ref{tab:results_iid}. The difference between \lxmerttext{} and \textsc{MajTempl} is small (3.3\%), suggesting that the answer to most questions cannot be inferred without looking at the images.  We also show that when the model is trained with random images instead of distracting ones (VB\textsubscript{\textsc{EasyDistractors}}), accuracy drops by 10.7\%, showing the importance of training on good distracting images. In addition, there is still a large gap from human performance, at 91.9\%, which we estimate by evaluating human answers on 160 questions.
Finally, we observe a 9.7\% performance drop when training on the automatically-generated examples and testing on the paraphrased examples. Accuracy per template is shown in App.~\ref{app:additional_results}.

Next, we report results on the compositional splits.
We show results on automatically-generated questions (not paraphrased), to disentangle the effect of compositional generalization from transfer to natural language.
App.~\ref{app:additional_results} reports results for the paraphrased test set, where generalization scores are lower, showing that transfer to natural language makes compositional generalization even harder.

Table~\ref{tab:results_few_shot} shows results in the \textbf{few-shot} setup, where in 5 out of 11 setups the generalization score is $\leq$ 70. VB\textsubscript{250} generalizes better in cases where the withheld operator is similar to an operator that appears in the training set. For instance, \textsc{Has-Quant-All} has higher generalization score compared to \textsc{Has-Quant} since it sees many examples with the quantifiers \emph{``some''} and \emph{``none''}, \textsc{Has-Compar-More} has a higher score compared to \textsc{Has-Compar}, and \textsc{Has-Logic-And} has a perfect generalization score. This suggests that when the model has some representation for a reasoning type it can generalize better to new instances of it.

The large gap between the nearly-perfect score of \textsc{Has-Num-3} (92\%), and the low score of \textsc{Has-Num-3-Ans-3} (0\%), where in both the number 3 is rarely seen in the question, and in the latter it is also rare as an answer, suggests that the model learns good number representations just from seeing numbers in the answers. Other cases where the generalization scores are low are \textsc{Has-Quant}, where quantifiers appear in only 250 examples, \textsc{Has-Quant-CompScope}, where
the scope of the quantifier is complex, and \textsc{Has-Compar}, where comparatives appear in only 250 examples.
Fig.~\ref{fig:m_effect} (App.~\ref{app:additional_results}) shows performance on the development set as $M$, the number of examples with the tested property that the model is shown during training, increases. We observe model performance is much lower when $M=50$ and improves rapidly as $M$ increases. This shows that models acquire new skills rapidly from hundreds of examples, but not from a handful of examples, like humans.

Table~\ref{tab:results_zero_shot} shows results for the \textbf{zero-shot} setup. A model that sees examples where the quantifier scope is complex, but never in the context of the quantifier \textsc{All}, fails to generalize (\textsc{Has-Quant-Comp \& Has-Quant-All}, 26\%). The model also fails to generalize to the template \textsc{ChooseObject}, although it saw at training time the necessary parts in the templates \textsc{ChooseAttr} and \textsc{VerifyObject}. Similarly, the model fails to generalize to the template \textsc{VerifyAttr},
and to \textsc{TPL-VerifyCount} $\cup$ \textsc{TPL-VerifyCountGroupBy}, where we hold out all verification questions with counting, even though the model sees verification questions and counting in other templates. Last, the model struggles to generalize in the program split.

Conversely, the model generalizes well to questions with the \texttt{Count} operator where the subgraph contains a complex sub-graph (\textsc{Has-Count \& RM/V/C}) or an attribute (\textsc{Has-Count \& Has-Attr}), and in the lexical split, where the model is tested on unseen combinations of names of nodes.

A possible explanation for the above is that compositional generalization is harder when the model needs to learn to combine large/complex structures, and performs better when composing more atomic constructs. However, further characterizing the conditions under which compositional generalization occurs is an important question for future work.

\section{Conclusion}
We present \visgroc{}, a test-bed for visually-grounded compositional generalization with real images. \visgroc{} is created automatically except for manual validation and paraphrasing, and allows us to create a suite of compositional splits. \visgroc{} can be easily extended with new templates and splits to encourage the community to further understand compositional generalization. Through \visgroc{}, we expose a wide range of cases where models struggle to compositionally generalize.

\section*{Acknowledgements}
This research was partially supported by The Yandex Initiative for Machine Learning, the European Research Council (ERC) under the European Union Horizons 2020 research and innovation programme (grant ERC DELPHI 802800), the DARPA MCS program under Contract No. N660011924033 with the United States Office Of Naval Research, and NSF award \#IIS-1817183. We thank the anonymous reviewers for their useful comments. This work was completed in partial fulfillment for the Ph.D degree of Ben Bogin.

\bibliography{anthology,custom}
\bibliographystyle{acl_natbib}

\clearpage

\appendix

\section{Filtering Overlapping Distracors}
\label{app:overlapping}

We take the published RoBERTa (Large, \citealt{liu2019roberta}) model that is already fine-tuned on MNLI \cite{williams-etal-2018-broad}, and further fine-tune it separately on pairs of nouns, attributes and relations to predict whether a pair of words or phrases are mutually exclusive. To leverage the knowledge learned during pre-training, we use the same setup as the training on MNLI, where the model is given two phrases and predicts one of three classes: ``contradiction'', ``entailment'' and ``neutral''.

To collect the list of pairs to annotate that will be used for fine-tuning, we fetched all pairs that have been used in Visual Genome within the same context. For \textbf{attributes}, we took all pairs of attributes that have appeared within the context of the same object (this way, we will be likely to collect \emph{``red''} and \emph{``green''} since they appear within the context objects such as \emph{``apple''}, but not \emph{``red''} and \emph{``grilled''}). For \textbf{nouns}, we consider all pairs of nouns that have been used with the same relation. For \textbf{relations}, we consider all pairs of relations that have been used with the same pair of nouns. While there are other resources that could have been useful for fine-tuning (e.g. WordNet, \citealt{wordnet}), we did not use any such external knowledge base since it allowed us to have exact control on the subtleties of the data in our training context.

We train all models for 50 epochs with a learning rate of $3e^{-5}$. For the \textbf{nouns} models, we use 2,366 manually annotated pairs of nouns for training and validation. The model is trained to predict ``contradiction'' whenever nouns are mutually-exclusive, i.e. when none of the words is a synonym, hypernym, or hyponym of the other, and ``neutral'' otherwise (we do not use the entailment class). We randomly shuffle the internal order of each pair for regularization. We get an accuracy score of 94.4\% on 20\% of the pairs which were held-out for validation. Similarly we train a model that predicts mutual-exclusiveness of \textbf{attributes} over 3,053 pairs, and get an accuracy of 95.7\%.

Unlike the other two models, for the \textbf{relations} model we do not require complete mutual-exclusiveness, and we do not assume symmetrical annotations, i.e., that if $m(x_1, x_2)$ then $m(x_2, x_1)$, to increase the probability of finding pairs where $m$ returns a score higher than 0.5 for a relation $x$. For example, we annotate pairs such that $m(\textrm{\emph{``riding on''}}, \textrm{\emph{``near''}})=1$ but $m(\textrm{\emph{``near''}}, \textrm{\emph{``'riding on'}})=0$, since \emph{most often}, if some object is hanging on another object, the annotation of the relations between the two objects in Visual Genmoe will be specific, i.e. \emph{``riding on''} or \emph{``on''} and not \emph{``near''}. This way, for a question such as \emph{``Is the man riding a motorcycle''} we might get distracting images with a man \emph{``standing near''} a motorcycle, but for a question such as \emph{``Is the man near a motorcycle''} we will not get distracting images with a man \emph{``riding''} a motorcycle, as then the question will be ambiguous. Note that while this can potentially introduce some noise (i.e., in some rare cases \emph{``a man riding a motorcycle''} might be annotated as if the man is \emph{``near''} a motorcycle), such mistakes will hopefully be overridden with the second validation that we use (incomplete scene graphs, App~\ref{app:incomplete}). We annotate 917 pairs of relations, where every pair is annotated in both directions. We get an accuracy of 82.5\% on the held-out set.

\section{Incomplete Scene Graphs}
\label{app:incomplete}

We use LXMERT \cite{tan-bansal-2019-lxmert} to train a classifier that predicts whether a simple subgraph exists in an image. See \S\ref{subsec:distracing} for details on the data we train on. We extract image and objects features with the bottom-up top-down attention method of \citet{anderson2018bottom} as performed in LXMERT's paper, and fine-tune the pre-trained model. To extract the training data, we use all subgraphs from all images for which we have at least one valid negative image (from both the training and test sets). This results in 6,520,367 positive and negative examples. Since we need the model to predict results not only on the test set, but also on the training set, we split all examples (training and test) into 5 splits based on their image, and train 5 different models, where each model does not see a different fifth of the images during training. Then, to predict whether a simple subgraph exists in an image, we use the model that was not trained on that image. 

We manually annotate 441 examples where we determine if a simple subgraph exists in an image and use these annotations for early stopping and to adjust a threshold $\tau$. We use this threshold to filter out candidate distracting images for a subgraph $g$ if the model outputs a score above a certain threshold $\tau$ for \emph{all} of the simple graphs in $g$.
Note that each negative example is a candidate distracting image to some subgraph $g$. We use $g$ to further adjust $\tau$ in the following way. By definition, a candidate simple graph of a distracting image has a non-empty set $d$ of nodes that are different than $g$. Based on our annotated examples, we found that the model should have a different threshold $\tau$ for different \emph{types} of nodes in $d$. Specifically, we found that the model performed best when $d$ contained nodes of type \texttt{object}, then \texttt{relation}, and finally \texttt{attribute}. Thus, we use a different $\tau$ for each type: $\tau=0.05$ for \texttt{object}, $\tau=0.1$ for \texttt{relation} and $\tau=0.5$ for \texttt{attribute}. If there are more than one type of nodes in $d$, we take the one that gives the maximal $\tau$.

The described procedure can be used to detect unannotated objects, however, it will not be useful in the non-rare case where an object is annotated in the scene graph, but the image contains more similar objects in the image (e.g. there is a crowd full of people in the image, but only a few of the people are annotated). We thus add another verification step for each simple sub-graph $g$. First, we take the annotated positions of all instances of $g$ in the scene. For example, if there are three annotated \textit{``apples''}, we will take the positions (bounding boxes annotations from Visual Genome) of all three. Then, we use our trained LXMERT model with the textual description of $g$ (e.g. \textit{``apple''}) and the image, but this time we zero-out the image parts that contain the apples according to their annotated positions.\footnote{LXMERT uses pre-calculated features of bounding boxes that are proposed by Faster-RCNN \cite{faster-rcnn}, thus we zero-out proposed bounding boxes that overlap with the annotated bounding boxes.} Essentially, we are querying the model if there are any other \textit{``apples''} other than those that are annotated. We use a similar procedure as before to find the best threshold, $0.5$. Since the LXMERT model is never trained with zeroed-out parts, during the described fine-tuning procedure we also zero-out 15\% of the bounding boxes.

\section{Downsampling \& Balancing}
\label{app:balance}
We use the following downsampling method to balance the dataset and reduce bias as much as possible, separately for the training, development and test sets. At a high-level, we start with a total of $N$ questions and group them by their templates, such that we have $T$ groups. We then use a heuristic ordering method that prioritizes or balances different desired features, described next, and finally we take the top $S=\frac{N}{T}$ questions from each group, such that we get an equal number of questions per template. The ordering method is defined as follows, starting with an empty list $L_t$ for a template $t$. Each question is automatically annotated with the following three features: (1) whether this question appears at least twice with different answers, (2) the answer to the question and (3) the structure of the source subgraph for that question, specifically a tuple with its size and its depth. We first add to $L_t$ all questions where the first feature is positive (in all cases this was less than $S$). Then, to balance between the different question answers, at each step until $|L_t|=S$, we count the appearances of all answers and sample an answer $a$ from the answers that appeared least in $L_t$. Then, we count the appearances of all subgraph structures, and sample a question with answer $a$, such that its subgraph structure appeared least. We stop once $|L_t|=S$.
\section{Additional Statistics}
\label{app:answers_dist}

\begin{figure}
  \centering
  \includegraphics[width=\linewidth]{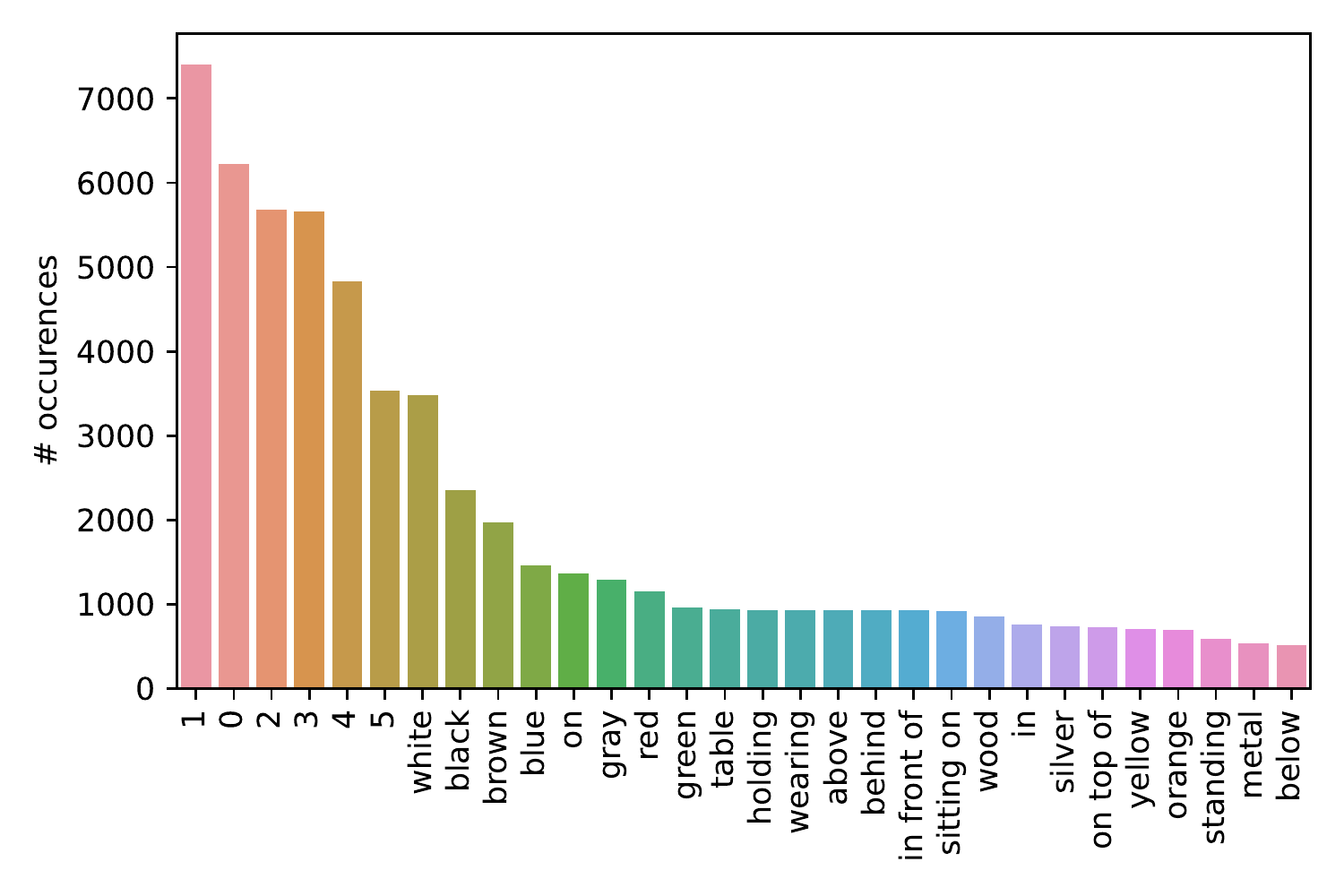}
  \caption{Distribution over the top 30 answers in the training set (excluding true/false).}
  \label{fig:answer_dist}
\end{figure}

\begin{figure}
  \centering
  \includegraphics[width=0.8\linewidth]{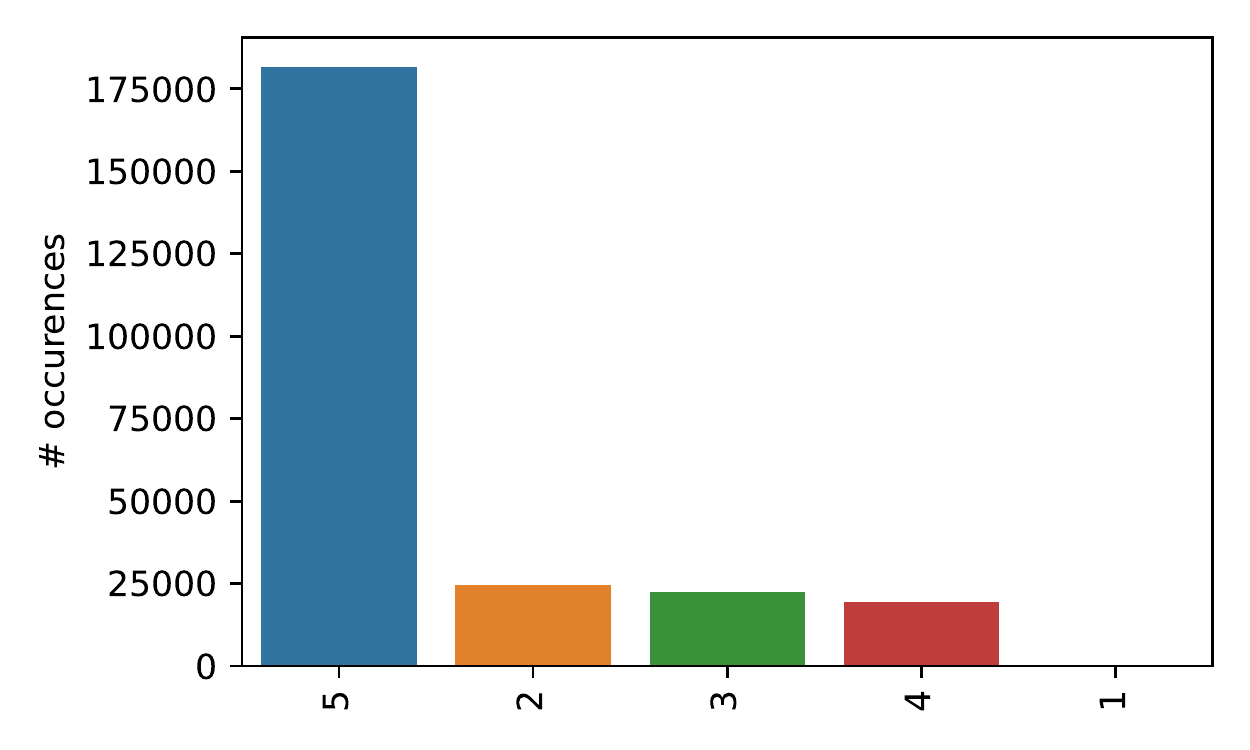}
  \caption{Distribution over the number of images each question in the training set contains.}
  \label{fig:num_images_dist}
\end{figure}

\begin{figure*}
  \centering
  \includegraphics[width=\linewidth]{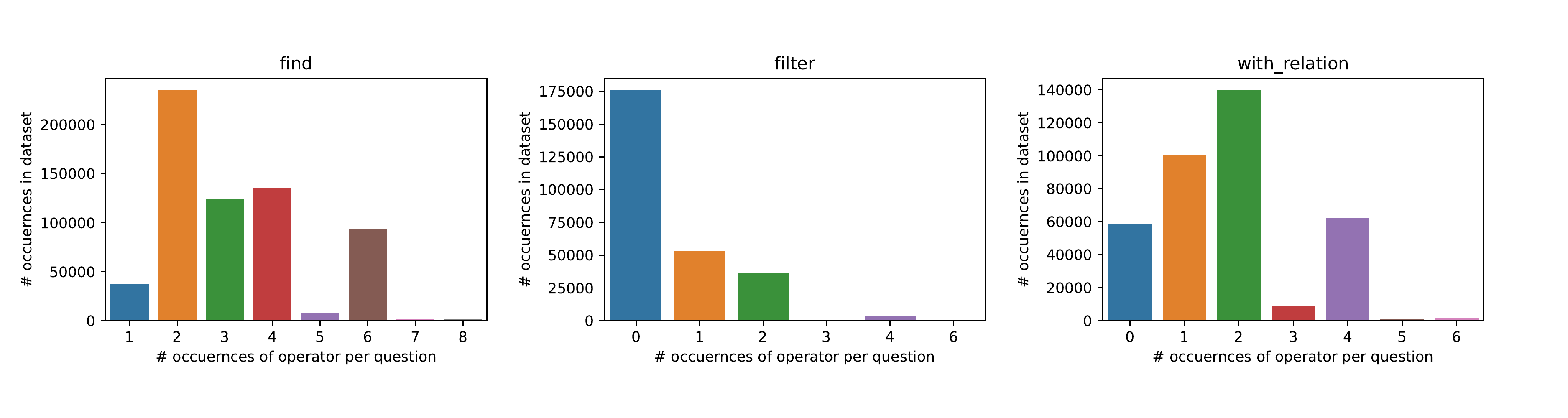}
  \caption{Distributions over the number of occurrences that specific operators appear in a single question in the training set.}
  \label{fig:num_operators_dist}
\end{figure*}

\paragraph{Answers distribution} We show the distribution over the top 30 answers of the training set in Fig.~\ref{fig:answer_dist}, excluding true/false answers. As can be seen, the most common answers are the numbers 0-5, followed by common colors, attributes and relations.

\paragraph{Number of images distribution} We show in Fig.~\ref{fig:num_images_dist} the distribution over the number of images each question in the training set contains. As can be seen, most questions contain exactly 5 images.

\paragraph{Occurrences of operators in questions} We show in Fig.~\ref{fig:num_operators_dist}, for a selected set of operators (\texttt{find}, \texttt{filter}, and \texttt{with\_relation}), the distribution over the number of occurrences of that operator in a single question (e.g. the program for the question \emph{``There is 1 green banana on a tree that is next to a man''} contains 3 \texttt{find} operators, one \texttt{filter} and two \texttt{with\_relation}). The graphs show that \texttt{find} appears between one to eight times in a single question, and \texttt{filter} and \texttt{with\_relation} between zero to six. Note that a question that contains six \texttt{with\_relation} does not imply that a single reference to an object contains six relations, since a question can contain more than one object reference (e.g. in \textsc{CompareCount}).

\section{Crowdsourcing Details}
\label{app:crowdsourcing}

We use Amazon Mechanical Turk (AMT) for two different tasks: validation of questions and paraphrasing them. 

\subsection{Validation}
\label{subapp:validation} We wanted to make sure that our validation and test examples are of high-quality by manually validating that the question is logically valid, there are no ambiguous object references, and the answer is correct. To maintain high-quality work in AMT, we first created a qualification task by annotating ourselves 100 examples, finding that the percentage of valid questions from the automatically generated samples was 83\%. Workers were asked to choose one of the following options: \emph{``Answer is CORRECT''}, \emph{``I cannot understand the question''}, \emph{``I cannot determine if the answer is correct''} or \emph{``Answer is WRONG''}. We filtered workers by their performance: workers that have gained over 85\% accuracy were given feedback and were approved for the main task that contained the rest of the questions. During their work, we have repeatedly sampled the annotations of the workers and gave feedback where necessary, and also measured the accuracy of their submissions: all workers got an accuracy of between 95\% to 98\%. Workers were paid 0.5\$ for a batch of 5 questions.

\begin{figure*}
  \centering
  \includegraphics[width=\linewidth]{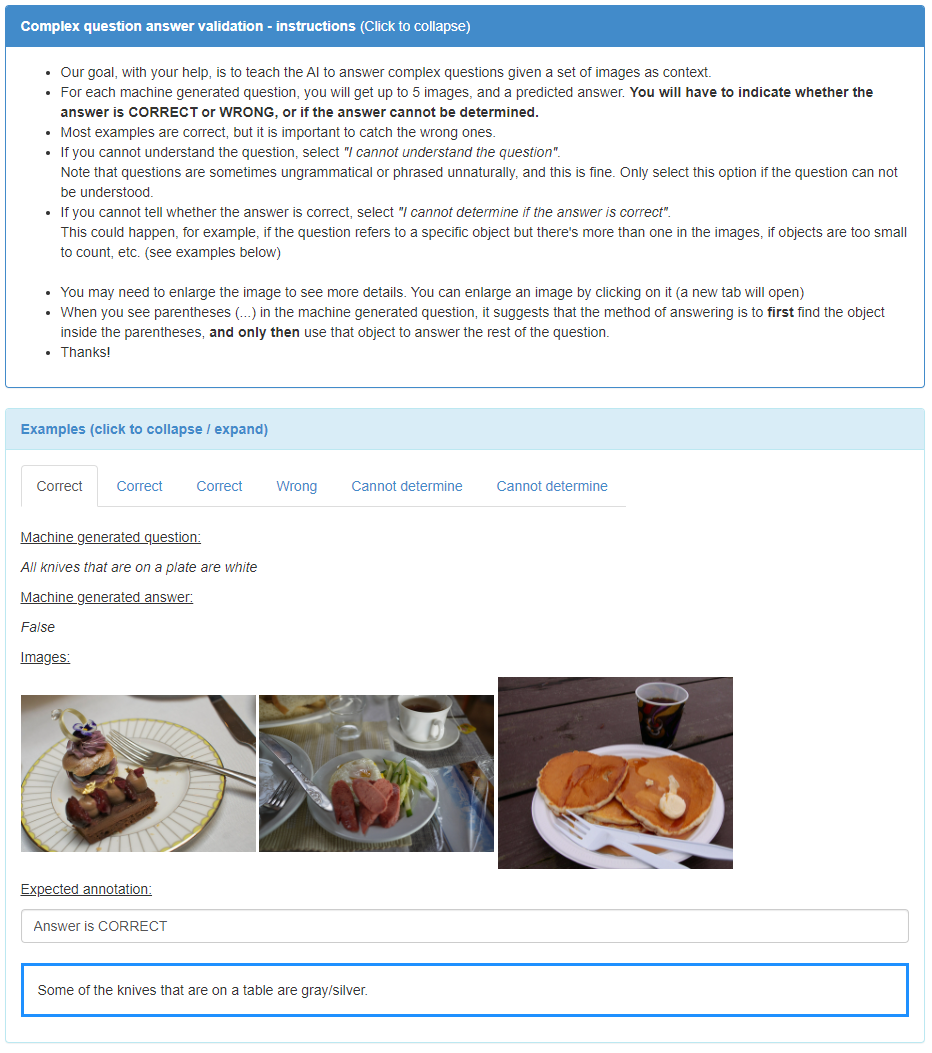}
  \caption{The instructions for the AMT validation task.}
  \label{fig:crowdsource_validation}
\end{figure*}

\begin{figure*}
  \centering
  \includegraphics[width=\linewidth]{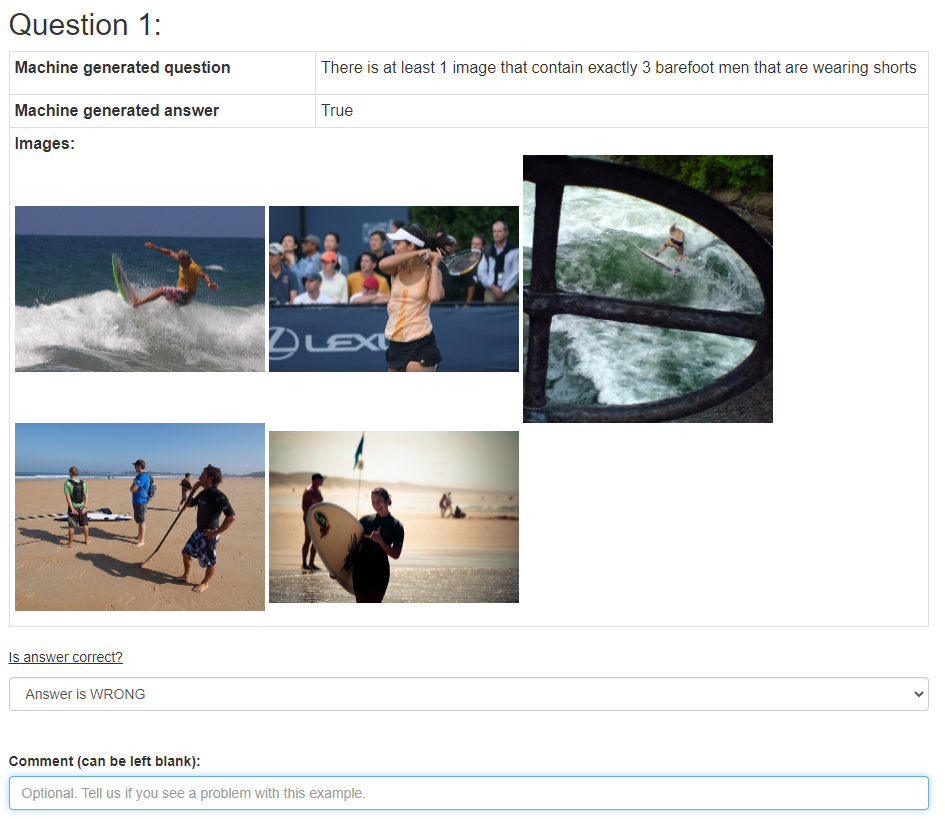}
  \caption{An example input for the AMT validation task. The expected annotation is ``Answer is WRONG''.}
  \label{fig:crowdsource_validation_hit}
\end{figure*}

We used parenthesis to clarify nested referral expressions, e.g. \emph{``How many pizzas are near (a silver fork that is next to a plate and is on a napkin )?''}. Screenshots of the instructions and the HIT can be seen in Figures~\ref{fig:crowdsource_validation} and \ref{fig:crowdsource_validation_hit}.

\paragraph{Analysis} We sample 40 examples that were filtered out by the annotators to analyze the different causes for invalid generated questions. We find that most errors (70\%) were due to problematic scene graph annotations: either because of missing annotations (53\%) or wrong annotations (17\%). The former type would make the answers to questions that require counting to be incorrect, and also questions that ask about a specific object (e.g. a question about \emph{``a man wearing a hat''} will be invalid if there's more than one such man), and in general, means that our automatic validation mechanism failed to recognize that object. The latter type (wrong annotations, in contrast to missing annotations) can cause any question to be incorrect, and could not have been detected by our automatic validation methods. Other errors were questions about color (6\%) that were not accurate (e.g. a question about whether two benches are brown will be given the answer `true' since they are both annotated brown, but in practice, they could have significantly different shades of brown which might lead to the correct answer 'false'). The rest of the errors (16\%) were due to various issues that make the answer unclear, such as questions that require to count feathers or meat.

\begin{figure*}
  \centering
  \includegraphics[width=\linewidth]{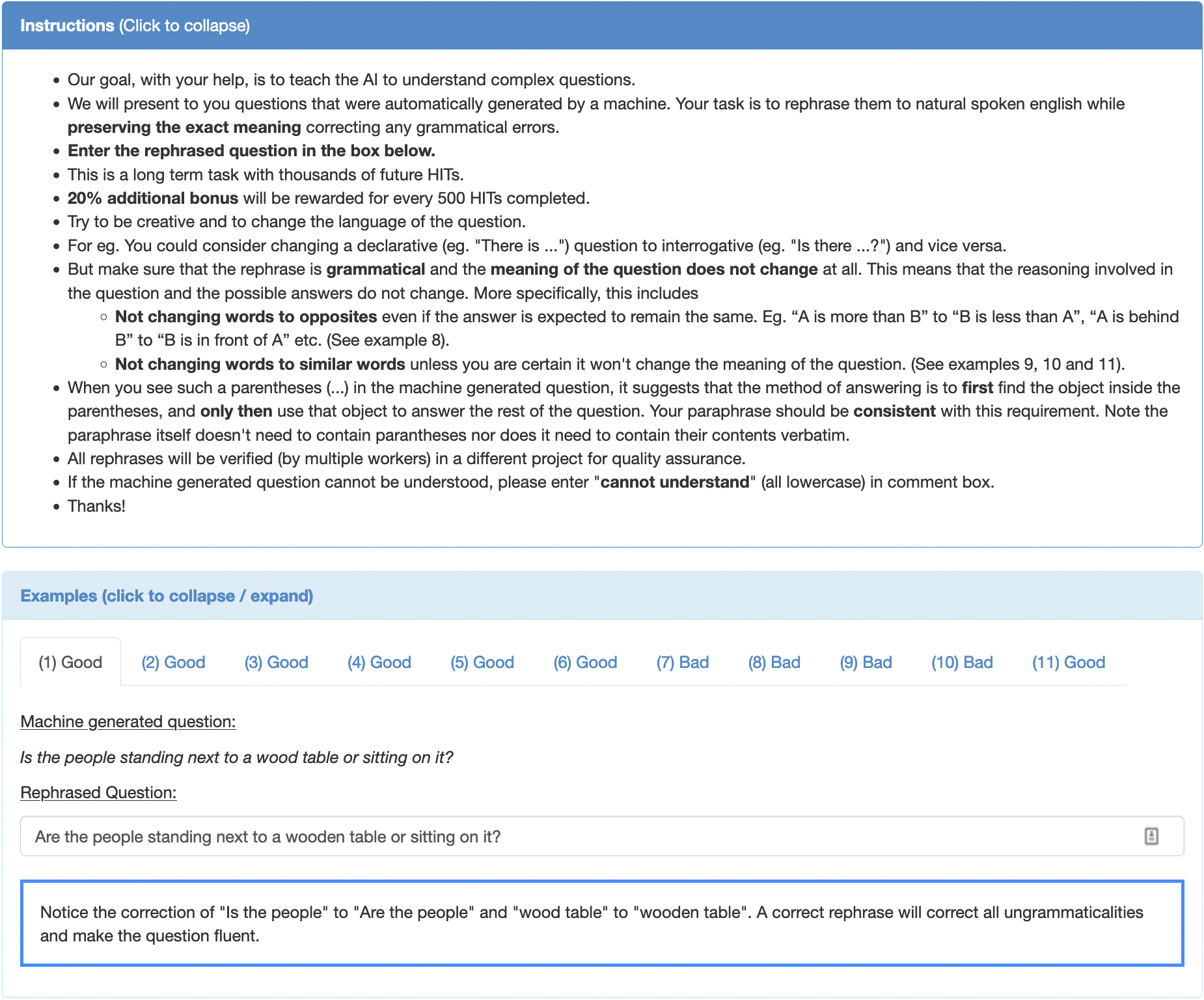}
  \caption{The instructions and examples for the AMT question rephrasing task.}
  \label{fig:crowdsourcing-rephrasing-instructions}
\end{figure*}

\begin{figure*}
  \centering
  \includegraphics[width=\linewidth]{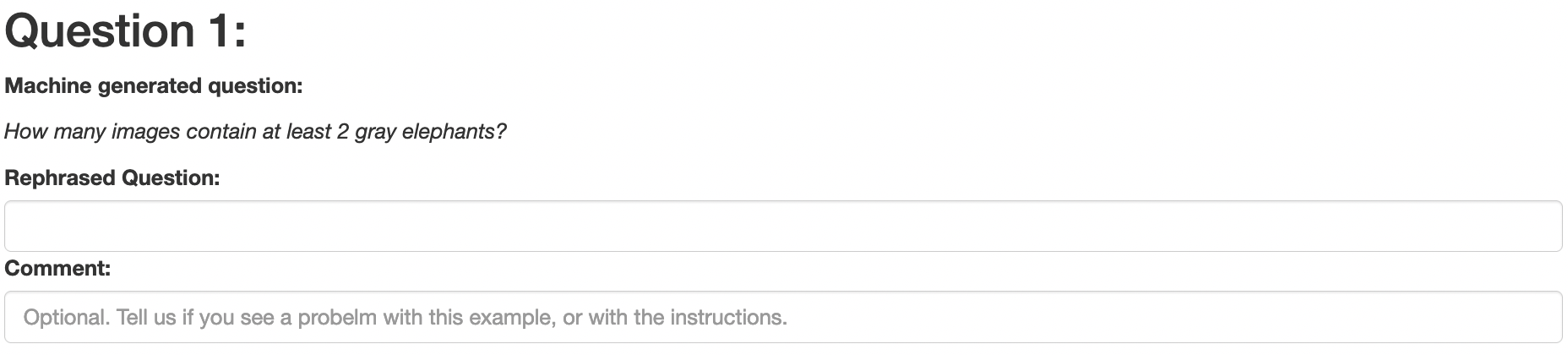}
  \caption{An example HIT for the AMT question rephrasing task.}
  \label{fig:crowdsourcing-rephrasing-HIT}
\end{figure*}

\begin{table*}
\centering
\small
\begin{tabular}{p{7.5cm}p{7.5cm}}
\toprule
                                                                                                                        Auto-generated question &                                                                                                  Paraphrase \\
    \midrule
                        Does the trees that are behind a zebra and the trees that are behind a fire hydrant have the same color? &                               Are the trees behind a zebra the same color as those behind a fire hydrant? \\
                                                                     There is 1 bottle that is on bench that is in front of tree &                                                          Is there a bottle on a bench in front of a tree? \\
                                                                                   No forks that are on a white plate are silver &                                                            None of the forks on a white plate are silver. \\
                                                                          Do all boats that are in a harbor have the same color? &                                                           Are all the boats in the harbor the same color? \\
                                                                           Is the person that is wearing a yellow jacket skiing? &                                                                Is the person in the yellow jacket skiing? \\
                                                          How many images with mushrooms that are on a pizza that is on a table? &                                                    The pizza on the table - how many mushrooms are on it? \\
    Is there either a girl that is holding a bouquet and is wearing dress or a girl that is holding a book and is wearing a hat? & Is there either a girl holding a bouquet and wearing a dress, or a girl holding a book and wearing a hat? \\
                                                        There are less boats that are on water than surfboards that are on water &                                                           There are fewer boats on water than surfboards. \\
                                                                              There are at least 4 people that are buttering pan &                                                            There are four or more people buttering a pan. \\
                                                                   What is the material of the table that is under a coffee mug? &                                                          What material is the table under the coffee mug? \\
    \bottomrule
\end{tabular}
\caption{Examples for crowd-sourced paraphrasing.}
\label{tab:paraphrase_samples}
\end{table*}

\subsection{Paraphrasing}
\label{subapp:paraphrasing}
For the question paraphrasing task, we again conducted a qualification task in addition to the final task. All potential workers were first added to the qualification task and asked to paraphrase 10 questions each. The paraphrases were then manually analyzed for meaning preservation and fluency and only the workers with very good performance were added to the final task which was used to paraphrase the bulk of the questions. In either case, we shared feedback with the workers via Google spreadsheets (one for each worker). Additionally, we regularly sampled and analysed the workers' paraphrases in the final task and used the same spreadsheets to share any necessary feedback. The workers were asked to periodically check their feedback spreadsheets and the workers that ignored the feedback were disqualified from the final task. We qualified 14 workers to the final task most of whom wrote good paraphrases. We only had to disqualify one worker for not taking note of their feedback. 

Both the qualification and final tasks had the same instructions, examples and HIT interface. Screenshots can be seen in Figures \ref{fig:crowdsourcing-rephrasing-instructions} and \ref{fig:crowdsourcing-rephrasing-HIT}. Workers were paid \$0.7 for every task completed in both AMT tasks -- with 5 questions per task. Additionally, as shown in Figure \ref{fig:crowdsourcing-rephrasing-HIT}, workers were provided a comment box to leave comments in case they could not understand the question. Comments were left for a very small fraction of questions (less than 2\%), mostly to indicate questions that were invalid or unclear. We removed all questions with comments in the final datasets. Some examples of the crowd-sourced paraphrases are shown in Table~\ref{tab:paraphrase_samples}.
\section{Dataset examples}
\label{app:examples}

\begin{figure*}
  \centering
  \includegraphics[width=0.95\linewidth]{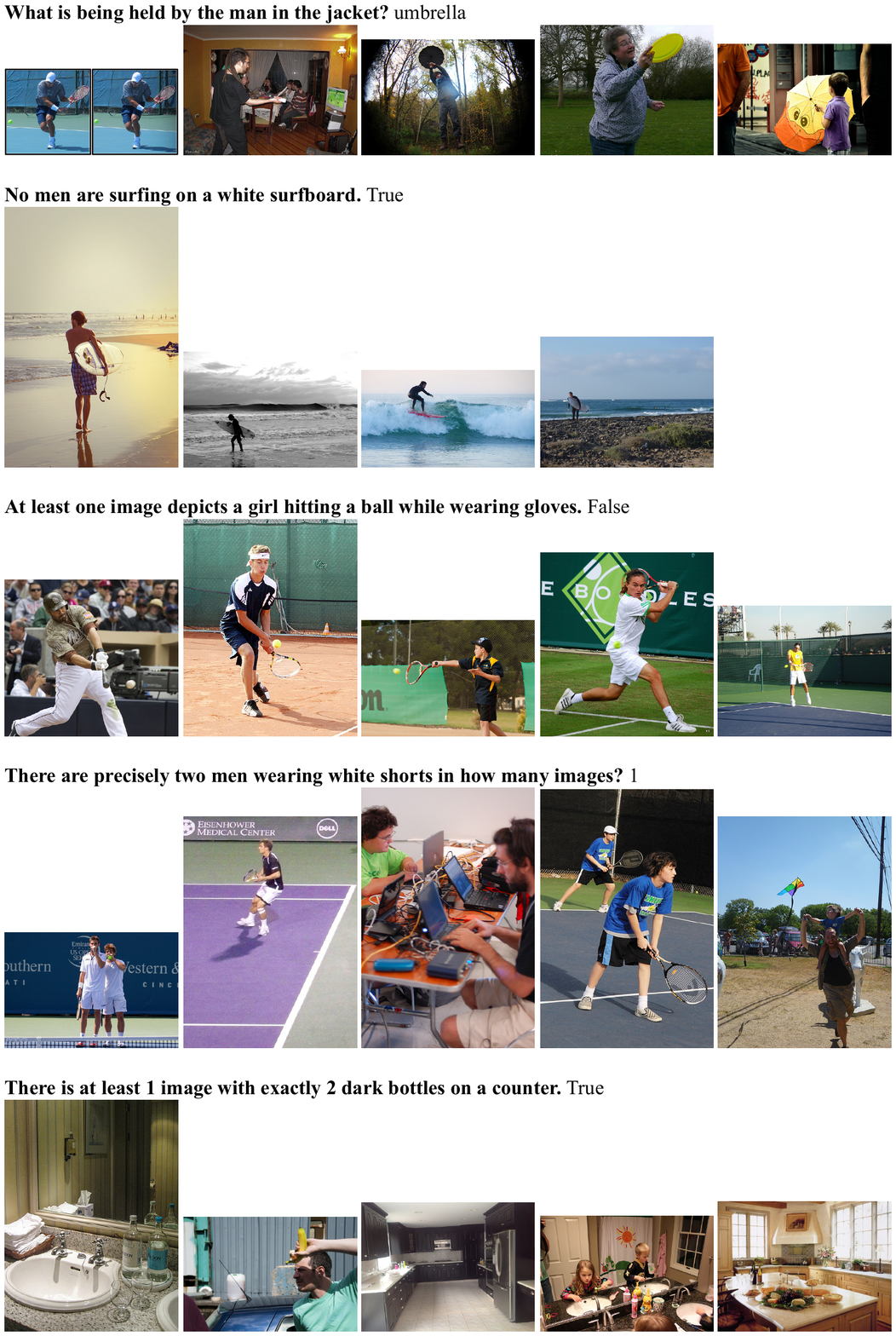}
  \caption{Selected examples from COVR validation set.}
  \label{fig:dataset_examples1}
\end{figure*}

\begin{figure*}
  \centering
  \includegraphics[width=0.95\linewidth]{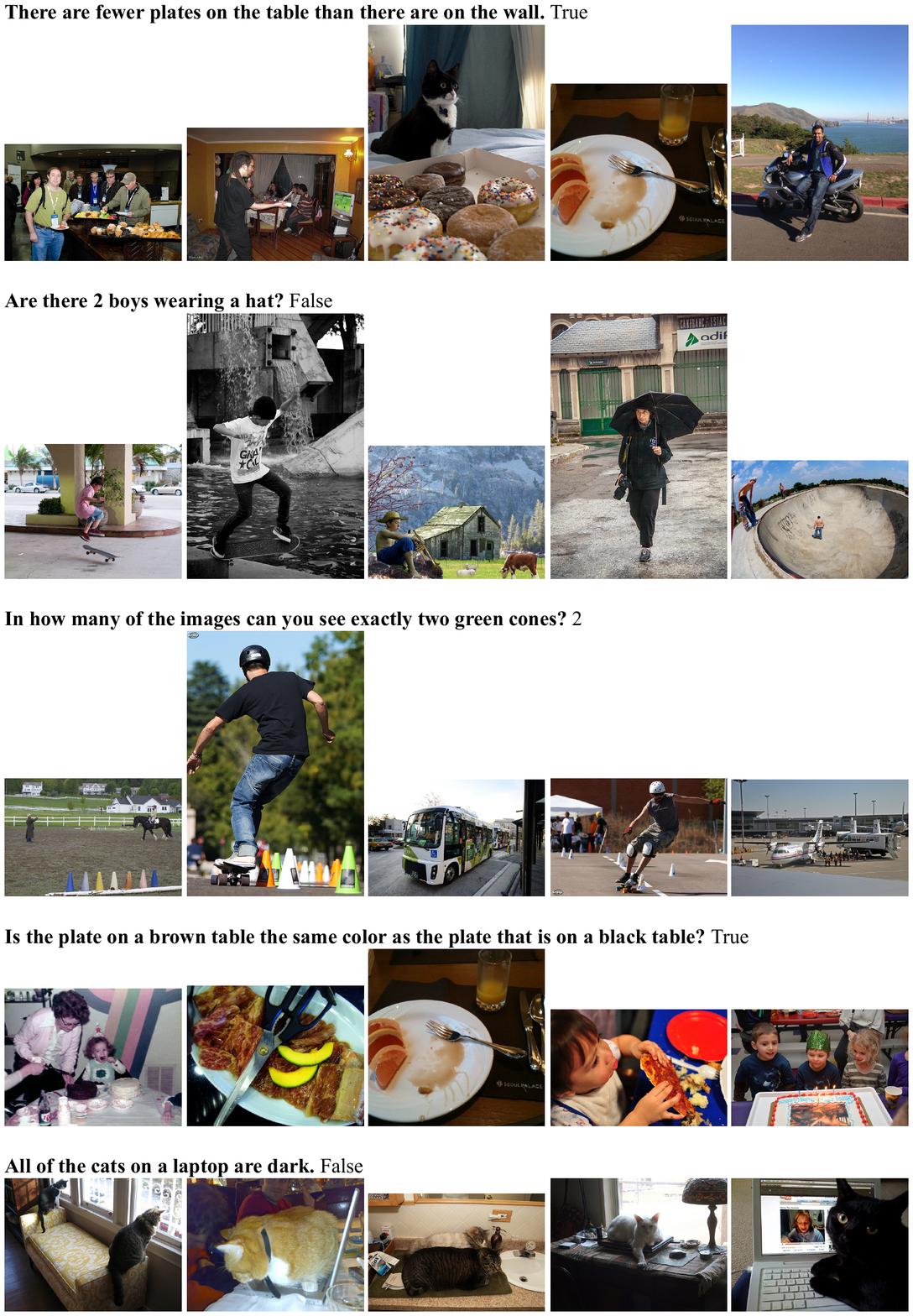}
  \caption{Selected examples from COVR validation set.}
  \label{fig:dataset_examples2}
\end{figure*}

Fig.~\ref{fig:dataset_examples1} and \ref{fig:dataset_examples2} show 10 selected validated and paraphrased examples from the validation set, demonstrating the variety of the questions and relevant distracting images.

\section{Programs}
\label{app:programs}

\begin{table}
\centering
\small
\begin{tabular}{lll}
\toprule
Index & Operator & Arguments \\
\midrule

 1 & \texttt{Find} & \emph{``table''} \\
 2 & \texttt{Filter} & $\Rsh$ 1, \emph{``wood''} \\
 3 & \texttt{Find} & \emph{``book''} \\
 4 & \texttt{WithRelation} & $\Rsh$ 3, $\Rsh$ 2, \emph{``on''} \\
 5 & \texttt{GroupByImages} & $\Rsh$ 4 \\
 6 & \texttt{KeepIfValuesCountEq} & $\Rsh$ 5, 2 \\
 7 & \texttt{Count} & $\Rsh$ 6 \\
  
 \bottomrule
\end{tabular}
\caption{The program for the question \emph{``How many images contain exactly 2 books that are on wood table?''}. The symbol $\Rsh$ with a row index next to it indicates that it is replaced with the output of the operator of the row in that index.}
\label{tab:sample_program}
\end{table}

\begin{table*}
\centering
\small
\begin{tabular}{p{3.75cm}p{3.5cm}p{8cm}}
\toprule
{\bf Operator} & {\bf Input} & {\bf Output} \\
 \midrule
\texttt{All} & (1) objects, (2) subprogram &  Returns `True' iff `subprogram' returns `True' for \emph{all} `objects'. \\
\texttt{Some} & (1) objects, (2) subprogram &  Returns `True' iff `subprogram' returns `True' for \emph{any} of the `objects'. \\
\texttt{None} & (1) objects, (2) subprogram &  Returns `True' iff `subprogram' returns `True' for \emph{none} of the objects. \\
\texttt{QueryName} & (1) object & Returns the name of `object' \\
\texttt{Find} & (1) name & Returns all objects from all scenes that are named `name'. \\
\texttt{Filter} & (1) objects (2) attribute\_value & Returns only objects in `objects' that have `attribute\_value'. \\
\texttt{Count} & (1) objects & Returns the size of `objects' \\
\texttt{Or} & (1) bool1 (2) bool2 & Returns `bool1' OR `bool2' \\
\texttt{And} & (1) bool1 (2) bool2 & Returns `bool1' AND `bool2' \\
\texttt{eq} & (1) number1 (2) number2 & Returns `True' iff number1 == number2 \\
\texttt{gt} & (1) number1 (2) number2 & Returns `True' iff number1 > number2 \\
\texttt{lt} & (1) number1 (2) number2 & Returns `True' iff number1 < number2 \\
\texttt{geq} & (1) number1 (2) number2 & Returns `True' iff number1 $\geq$ number2 \\
\texttt{leq} & (1) number1 (2) number2 & Returns `True' iff number1 $\leq$ number2 \\
\texttt{Unique} & (1) objects & Assumes `objects' contain a single object. Returns the object in the list. \\
\texttt{UniqueImages} & (1) objects & Returns a set (without duplicates) of all images of the given `objects'. \\
\texttt{GroupByImages} & (1) objects & Returns (image, objects\_in\_image) tuples where all object in `objects' that are in the same image are grouped together and coupled with that image. \\
\texttt{KeepIfValuesCountEq}/\newline\texttt{KeepIfValuesCountGt}/\newline\texttt{KeepIfValuesCountLt} & (1) (key, list) tuples \newline (2) size & Returns only tuples where the size of `list' is equal/greater than/less than `size'. \\
\texttt{QueryAttribute} & (1) object \newline (2) attribute\_name & Returns the attribute value (e.g., \emph{``red''}) of the `attribute\_name' (e.g., \emph{``color''}) of `object'. \\
\texttt{VerifyAttribute} & (1) object \newline (2) attribute\_value & Returns ``True'' iff `object' has the attribute `attribute\_value'. \\
\texttt{WithRelation} & (1) objects1, (2) objects2 \newline (3) relation & Returns all objects from `objects1` that have the relation `relation' with any of the objects in `objects2. \\
\texttt{WithRelationObject} & (1) objects1, (2) objects2 \newline (3) relation & Same as \texttt{WithRelation}, except it returns objects from `objects2'. \\
  
 \bottomrule
\end{tabular}
\caption{All program operators.}
\label{tab:operators}
\end{table*}

We list all program operators in Table~\ref{tab:operators}, together with their input arguments/dependencies and output. A sample program can be found in Table~\ref{tab:sample_program}.
\section{Additional Results}
\label{app:additional_results}

\begin{table}
\centering
\small
\begin{tabular}{lll}
\toprule
{\bf Template} & {\bf \lxmerttext{}} & {\bf \lxmertiid{}} \\
 \midrule
  \textsc{VerifyAttr} & 49.6 & 73.5 \\
  \textsc{ChooseAttr} & 52.0 & 69.5 \\
  \textsc{QueryAttr} & 36.2 & 47.1 \\
  \textsc{CompareCount} & 54.3 & 80.0 \\
  \textsc{Count} & 26.2 & 73.1 \\
  \textsc{VerifyCount} & 50.0 & 87.7 \\
  \textsc{CountGroupBy} & 26.6 & 60.3 \\
  \textsc{VerifyCount-GroupBy} & 49.5 & 71.2 \\
  \textsc{VerifyLogic} & 50.4 & 77.5 \\
  \textsc{VerifyQuantifier} & 51.2 & 80.7 \\
  \textsc{VerifyQuantifier-}\textsc{Attr} & 50.4 & 80.3 \\
  \textsc{ChooseObject} & 52.0 & 66.4 \\
  \textsc{QueryObject} & 11.0 & 14.4 \\
  \textsc{VerifySameAttr} & 52.1 & 62.3 \\
  \textsc{ChooseRel} & 58.1 & 65.9 \\
 \bottomrule
\end{tabular}
\caption{Accuracy score per template (i.i.d setup) on \visgroc{} automatically-generated questions, test set.}
\label{tab:accuracy_per_template}
\end{table}

\begin{table}
\centering
\small
\begin{tabular}{lll}
\toprule
{\bf Template} & {\bf \lxmerttext{}} & {\bf \lxmertiid{}} \\
 \midrule
  \textsc{VerifyAttr} & 46.5 & 69.0 \\
  \textsc{ChooseAttr} & 48.5 & 62.1 \\
  \textsc{QueryAttr} & 23.3 & 35.2 \\
  \textsc{CompareCount} & 50.9 & 67.8\\
  \textsc{Count} & 25.2 & 67.6 \\
  \textsc{VerifyCount} & 50.2 & 80.9 \\
  \textsc{CountGroupBy} & 29.6 & 46.0 \\
  \textsc{VerifyCount-GroupBy} & 50.0 & 67.1 \\
  \textsc{VerifyLogic} & 47.7 & 71.1 \\
  \textsc{VerifyQuantifier} & 50.0 & 68.7 \\
  \textsc{VerifyQuantifier-}\textsc{Attr} & 50.0 &  75.6\\
  \textsc{ChooseObject} & 19.6 & 41.9 \\
  \textsc{QueryObject} & 6.3 & 8.3 \\
  \textsc{VerifySameAttr} & 47.9 & 50.7 \\
  \textsc{ChooseRel} & 51.8 & 52.2 \\
 \bottomrule
\end{tabular}
\caption{Accuracy score per template (i.i.d setup) on \visgroc{} (paraphrased), test set.}
\label{tab:accuracy_per_template_paraph}
\end{table}

\begin{table*}
\centering
\scriptsize
\begin{tabular}{lllllll}
\toprule
Split & Filtered & \lxmerttext{} & VB\textsubscript{250} & Gen. Score & VB\textsubscript{iid-size} & VB\textsubscript{iid} \\ \midrule
\textsc{Has-Quant} & 33.3k & 50.0 & 48.8 & \progressbar[linecolor=red]{0} & 70.9 & 72.1  \\
\textsc{Has-Quant-All} & 21.1k & 50.2 & 58.7 & \progressbar{0.43} & 70.1 & 74.1  \\
\textsc{Has-Quant-CompScope} & 22.8k & 49.8 & 53.8 & \progressbar{0.19} & 70.3 & 72.5  \\
\textsc{Has-Compar} & 16.7k & 50.9 & 44.8 & \progressbar[linecolor=red]{0} & 64.8 & 67.8  \\
\textsc{Has-Compar-More} & 7.6k & 50.8 & 60.7 & \progressbar{0.5} & 70.6 & 72.9  \\
\textsc{Has-GroupBy} & 33.3k & 39.8 & 54.0 & \progressbar{0.79} & 57.8 & 56.6  \\
\textsc{Has-Logic} & 16.7k & 47.7 & 55.7 & \progressbar{0.33} & 71.9 & 71.1  \\
\textsc{Has-Logic-And} & 8.8k & 49.6 & 72.7 & \progressbar{0.97} & 73.4 & 73.0  \\
\textsc{Has-Num-3} & 6.9k & 42.7 & 65.6 & \progressbar{1.1} & 63.5 & 63.5  \\
\textsc{Has-Num-3-Ans-3} & 12.1k & 24.2 & 25.4 & \progressbar{0.04} & 54.5 & 51.6  \\
\textsc{Ans-Num} & 33.3k & 27.3 & 28.8 & \progressbar{0.05} & 57.9 & 57.3  \\
\bottomrule
\end{tabular}
\caption{Same splits and experiments as in Table~\ref{tab:results_few_shot}, evaluated on the paraphrased questions.}
\label{tab:results_few_shot_para}
\end{table*}

\begin{table*}
\centering
\scriptsize
\begin{tabular}{m{6cm}lllllll}
\toprule
Split & Filtered & \lxmerttext{} & VB\textsubscript{0} & Gen. Score & VB\textsubscript{iid-size} & VB\textsubscript{iid} \\ \midrule
\textsc{Has-Quant-Comp \& Has-Quant-All} & 12.9k & 50.1 & 54.5 & \progressbar{0.24} & 68.9 & 75.0  \\
\textsc{Has-Count \& Has-Attr} & 37.6k & 40.9 & 52.7 & \progressbar{0.98} & 53.0 & 55.7  \\
\textsc{Has-Count \& RM/V/C} & 36.9k & 39.9 & 68.5 & \progressbar{0.79} & 76.3 & 76.5  \\
\textsc{Has-SameAttr-Color} & 27.3k & 47.7 & 59.3 & \progressbar{0.84} & 61.6 & 63.5  \\
\textsc{TPL-ChooseObject} & 16.7k & 19.6 & 1.6 & \progressbar[linecolor=red]{0} & 34.8 & 41.9  \\
\textsc{TPL-VerifyQuantAttr} & 16.7k & 50.0 & 57.8 & \progressbar{0.36} & 72.0 & 75.6  \\
\textsc{TPL-VerifyAttr} & 16.7k & 46.5 & 8.1 & \progressbar[linecolor=red]{0} & 64.0 & 69.0  \\
\textsc{TPL-VerifyCount} $\cup$ \textsc{TPL-VerifyCountGroupBy} & 33.3k & 50.1 & 47.7 & \progressbar[linecolor=red]{0} & 73.7 & 74.3  \\
\midrule
Program Split & 48k $\pm$ 11k & 38.6 $\pm$ 2.5 & 42.8 $\pm$ 3.8 & \progressbar{0.31} & 52.3 $\pm$ 2.0 & 55.5 $\pm$ 2.7  \\
Lexical Split & 40k $\pm$ 3k & 43.1 $\pm$ 0.6 & 60.2 $\pm$ 0.8 & \progressbar{0.88} & 62.6 $\pm$ 0.9 & 64.4 $\pm$ 0.8  \\
\bottomrule
\end{tabular}
\caption{Same splits and experiments as in Table~\ref{tab:results_zero_shot}, evaluated on the paraphrased questions.}
\label{tab:results_zero_shot_para}
\end{table*}

\begin{figure}
  \centering
  \includegraphics[width=0.95\linewidth]{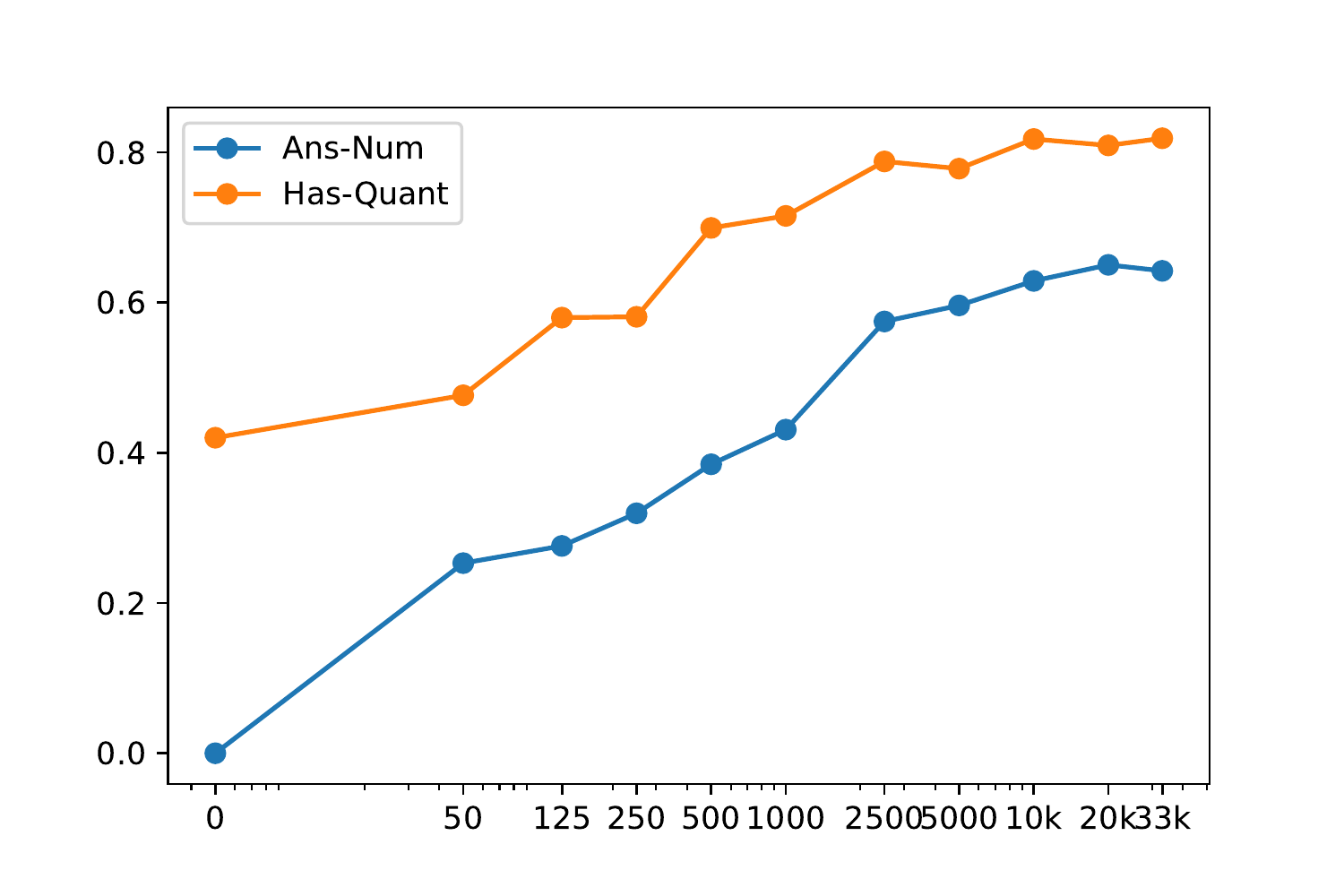}
  \caption{Effect of $M$ on development set accuracy on two compositional splits, starting with no examples at all of the tested compositional property (zero-shot) and up to the maximal amount of available training examples with that property. Log scale is used for the X-axis.}
  \label{fig:m_effect}
\end{figure}

\begin{table*}
\centering
\scriptsize
\begin{tabular}{lllllll}
\toprule
Split & Filtered & VLB\textsubscript{\textsc{Text}} & VLB\textsubscript{250} & Gen. Score & VLB\textsubscript{iid-size} & VLB\textsubscript{iid} \\ \midrule
\textsc{Has-Quant} & 33.3k & 51.1 & 65.6 & \progressbar{0.53} & 78.5 & 80.7  \\
\textsc{Has-Quant-All} & 21.1k & 50.9 & 60.9 & \progressbar{0.43} & 74.1 & 77.6  \\
\textsc{Has-Quant-CompScope} & 22.8k & 50.9 & 67.5 & \progressbar{0.61} & 78.1 & 80.2  \\
\textsc{Has-Compar} & 16.7k & 54.3 & 64.2 & \progressbar{0.44} & 76.8 & 77.7  \\
\textsc{Has-Compar-More} & 7.6k & 53.8 & 79.8 & \progressbar{1.03} & 79.0 & 81.7  \\
\textsc{Has-GroupBy} & 33.3k & 37.2 & 55.6 & \progressbar{0.7} & 63.4 & 61.3  \\
\textsc{Has-Logic} & 16.7k & 50.2 & 72.1 & \progressbar{0.84} & 76.2 & 75.0  \\
\textsc{Has-Logic-And} & 8.8k & 50.0 & 70.9 & \progressbar{0.82} & 75.5 & 74.5  \\
\textsc{Has-Num-3} & 6.9k & 43.8 & 67.7 & \progressbar{0.96} & 68.8 & 71.9  \\
\textsc{Has-Num-3-Ans-3} & 12.1k & 25.4 & 27.5 & \progressbar{0.05} & 66.4 & 68.0  \\
\textsc{Ans-Num} & 33.3k & 24.1 & 38.5 & \progressbar{0.36} & 64.0 & 65.8  \\
\bottomrule
\end{tabular}
\caption{Same splits and experiments as in Table~\ref{tab:results_few_shot}, for \textsc{ViLBERT}.}
\label{tab:results_few_shot_vilbert}
\end{table*}

\begin{table*}
\centering
\scriptsize
\begin{tabular}{m{6cm}lllllll}
\toprule
Split & Filtered & VLB\textsubscript{\textsc{Text}} & VLB\textsubscript{0} & Gen. Score & VLB\textsubscript{iid-size} & VLB\textsubscript{iid} \\ \midrule
\textsc{Has-Quant-CompScope \& Has-Quant-All} & 12.9k & 50.8 & 64.7 & \progressbar{0.54} & 76.4 & 77.4  \\
\textsc{Has-Count \& Has-Attr} & 37.6k & 39.6 & 57.4 & \progressbar{0.77} & 62.6 & 62.6  \\
\textsc{Has-Count \& RM/V/C} & 36.9k & 39.9 & 71.4 & \progressbar{0.76} & 81.1 & 81.9  \\
\textsc{Has-SameAttr-Color} & 27.3k & 48.5 & 64.7 & \progressbar{0.84} & 67.7 & 71.4  \\
\textsc{TPL-ChooseObject} & 16.7k & 51.0 & 2.0 & \progressbar[linecolor=red]{0} & 58.9 & 63.8  \\
\textsc{TPL-VerifyQuantAttr} & 16.7k & 50.4 & 61.2 & \progressbar{0.42} & 76.1 & 78.2  \\
\textsc{TPL-VerifyAttr} & 16.7k & 50.2 & 0.0 & \progressbar[linecolor=red]{0} & 70.0 & 75.4  \\
\textsc{TPL-VerifyCount} $\cup$ \textsc{TPL-VerifyCountGroupBy} & 33.3k & 49.6 & 29.5 & \progressbar[linecolor=red]{0} & 78.0 & 77.1  \\
\midrule
Program Split & 48k $\pm$ 11k & 43.6 $\pm$ 4.5 & 49.0 $\pm$ 3.1 & \progressbar{0.29} & 61.9 $\pm$ 4.5 & 64.6 $\pm$ 4.9  \\
Lexical Split & 40k $\pm$ 3k & 46.4 $\pm$ 0.4 & 70.4 $\pm$ 1.1 & \progressbar{0.95} & 71.7 $\pm$ 0.2 & 73.7 $\pm$ 0.5  \\
\bottomrule
\end{tabular}
\caption{Same splits and experiments as in Table~\ref{tab:results_zero_shot}, for \textsc{ViLBERT}.}
\label{tab:results_zero_shot_vilbert}
\end{table*}

\paragraph{Results per pattern} Tables~\ref{tab:accuracy_per_template} and \ref{tab:accuracy_per_template_paraph} show the accuracy for each template for both the non-paraphrased and the paraphrased versions, for models that were trained on all data. The results of the text-only baseline, \lxmerttext{}, show that indeed the model is struggling to get more than a random baseline performance of 50\% in most patterns. The scores of the main model, \lxmertiid{}, show that open questions are hardest (\textsc{QueryAttr}, \textsc{QueryObject}), and that there's a rather large variance between the performance on the different patterns.

\paragraph{Compositional Results on COVR-\textsc{Paraph.}}

We report results on the compositional splits when they are evaluated on the \emph{paraphrased} questions in Tables~\ref{tab:results_few_shot_para} and \ref{tab:results_zero_shot_para}. The generalization scores are lower than the results for the non-paraphrased data, showing that transfer to natural language makes compositional generalization harder.

\paragraph{Effect of $M$}

We show how $M$, the number of examples the model sees from the compositional subset, affects the accuracy in Figure~\ref{fig:m_effect}. The graph shows that using 50 examples barely has an effect, and that most of the improvement is achieved when increasing the number of examples from 125 to 2500. Increasing it further shows diminishing improvements.

\paragraph{\textsc{ViLBERT} Results}

To assess whether the results we get are specific to the model that we used (VisualBERT), we run additional compositional tests on a different model, \textsc{ViLBERT}, using Volta's framework \cite{bugliarello-etal-2021-multimodal}. The model has the same number of parameters and was trained on the same pre-training data. Results in Tables~\ref{tab:results_few_shot_vilbert} and \ref{tab:results_zero_shot_vilbert} show that for most of the compositional splits, both of our tested models get similar generalization scores.

\end{document}